\documentclass[letterpaper, 10 pt, conference]{ieeeconf}  % Comment this line out if you need a4paper

\IEEEoverridecommandlockouts                              % This command is only needed if 
\usepackage{multicol}

\usepackage{amsmath}
\usepackage{amssymb}
\usepackage{color}
\usepackage[usenames,dvipsnames]{xcolor}
\usepackage{graphicx}
\usepackage[font=small]{caption}
\usepackage[position=b]{subcaption}
\usepackage{multirow}
\usepackage{siunitx}
\usepackage{balance}
\usepackage{MnSymbol}
\usepackage{cite}
\usepackage{bbm}

\usepackage{algorithm}
\usepackage{algpseudocode}

\usepackage{tikz}
\usetikzlibrary{automata,arrows,shapes,shapes.geometric,matrix}
\usetikzlibrary{fit}

\newcommand{\real}{\mathbb{R}}
\newcommand{\distr}{\mathcal{D}}
\newcommand{\pdist}{\mathcal{P}_\mathbf{x}}
\newcommand{\expv}{\mathbb{E}}

\newcommand{\reach}{\mathbb{P}}

\newcommand{\on}{\mathsf{on}}
\newcommand{\off}{\mathsf{off}}
\newcommand{\start}{\mathsf{start}}
\newcommand{\boot}{\mathsf{boot}}
\newcommand{\sbo}{\mathsf{sbo}}

\newcommand{\N}{\mathcal{N}}
\newcommand{\bigO}{\mathcal{O}}

\newcommand{\Bz}{\mathbf{\mathit{b}}}
\newcommand{\B}{\mathbf{b}}
\newcommand{\Bspace}{\mathbb{B}}
\newcommand{\Bof}{\bar{\B}}
\newcommand{\traj}{\varphi}
\newcommand{\sched}{\varsigma}
\newcommand{\trigger}{\xi}

\newcommand{\coll}{\mathsf{coll}}
\newcommand{\targ}{\mathsf{targ}}
\newcommand{\free}{\mathsf{free}}
\newcommand{\cost}{\mathsf{cost}}

\newcommand{\vx}{\mathbf{x}}
\newcommand{\vz}{\mathbf{z}}
\newcommand{\vv}{\mathbf{v}}
\newcommand{\vu}{\mathbf{u}}
\newcommand{\vw}{\mathbf{w}}
\newcommand{\veps}{\boldsymbol{\epsilon}}
\newcommand{\mQ}{\mathbf{Q}_w}
\newcommand{\mR}{\mathbf{Q}_v}
\newcommand{\loc}{a}
\newcommand{\cov}{\mathbf{Q}}

\newcommand{\mdp}{\mathcal{M}}
\newcommand{\init}{\mathsf{init}}

\newcommand{\mdps}{S}
\newcommand{\mdpa}{Act}
\newcommand{\mdpp}{P}
\newcommand{\mdpc}{C}
\newcommand{\od}{\mathsf{od}}
\newcommand{\lo}{\mathsf{lo}}
\newcommand{\ie}{{\it i.e., }}
\newcommand{\eg}{{\it e.g., }}
\newcommand{\mean}{\mathsf{mean}}
\newcommand{\var}{\mathsf{var}}

\newtheorem{problem}{Problem}

\title{\LARGE \bf
Resource-Performance Trade-off Analysis for Mobile Robot Design
}

\author{M. Lahijanian, M. Svorenova, A. A. Morye, B. Yeomans, D. Rao, I. Posner, \\ P. Newman, H. Kress-Gazit, M. Kwiatkowska
}

\begin{document}

\maketitle
\thispagestyle{plain} %empty
\pagestyle{plain} %empty

%%%%%%%%%%%%%%%%%%%%%%%%%%%%%%%%%%%%%%%%%%%%%%%%%%%%%%%%%%%%%%%%%%%%%%%%%%%%%%%%
\begin{abstract} \label{sec:abstract}
%Abstract goes here ...
The design of mobile autonomous robots is challenging due to the limited on-board resources such as processing power and energy. A promising approach is to generate intelligent schedules
that reduce the resource consumption while maintaining best performance, or more interestingly, to trade off reduced resource consumption for a slightly lower but still acceptable level of performance. In this paper, we provide a framework to aid designers in exploring such resource-performance trade-offs and finding schedules for mobile robots, guided by questions such as ``what is the minimum resource budget required to achieve a given level of performance?'' The framework is based on a quantitative multi-objective verification technique which, for a collection of possibly conflicting objectives, produces the Pareto front that contains all the optimal trade-offs that are achievable. The designer then selects a specific Pareto point based on the resource constraints and desired performance level, and a correct-by-construction schedule that meets those constraints is automatically generated. We demonstrate the efficacy of this framework on several robotic scenarios in both simulations and experiments with encouraging results.
\end{abstract}

\section{Introduction}\label{sec:intro}
Mobile robotics is a fast growing field with a broad range of applications such as home appliance, aerial vehicles, and space exploration. The main feature that makes these robots very attractive from the application perspective is their ability to operate remotely with some level of autonomy.  The very same factors, however, introduce a challenge from the design angle due to the limited on-board resources such as processing power and energy source. For example, the Curiosity Mars rover operates on a CPU with less computational power than a today's typical smartphone CPU~\cite{rad750_cpu_specs}, resulting in slow movements and limited capabilities of the rover. In drones, the weight and the capacity of the on-board battery directly influences the robot's ability to stay airborne. 

\begin{figure}
\centering{
\includegraphics[width=1\columnwidth]{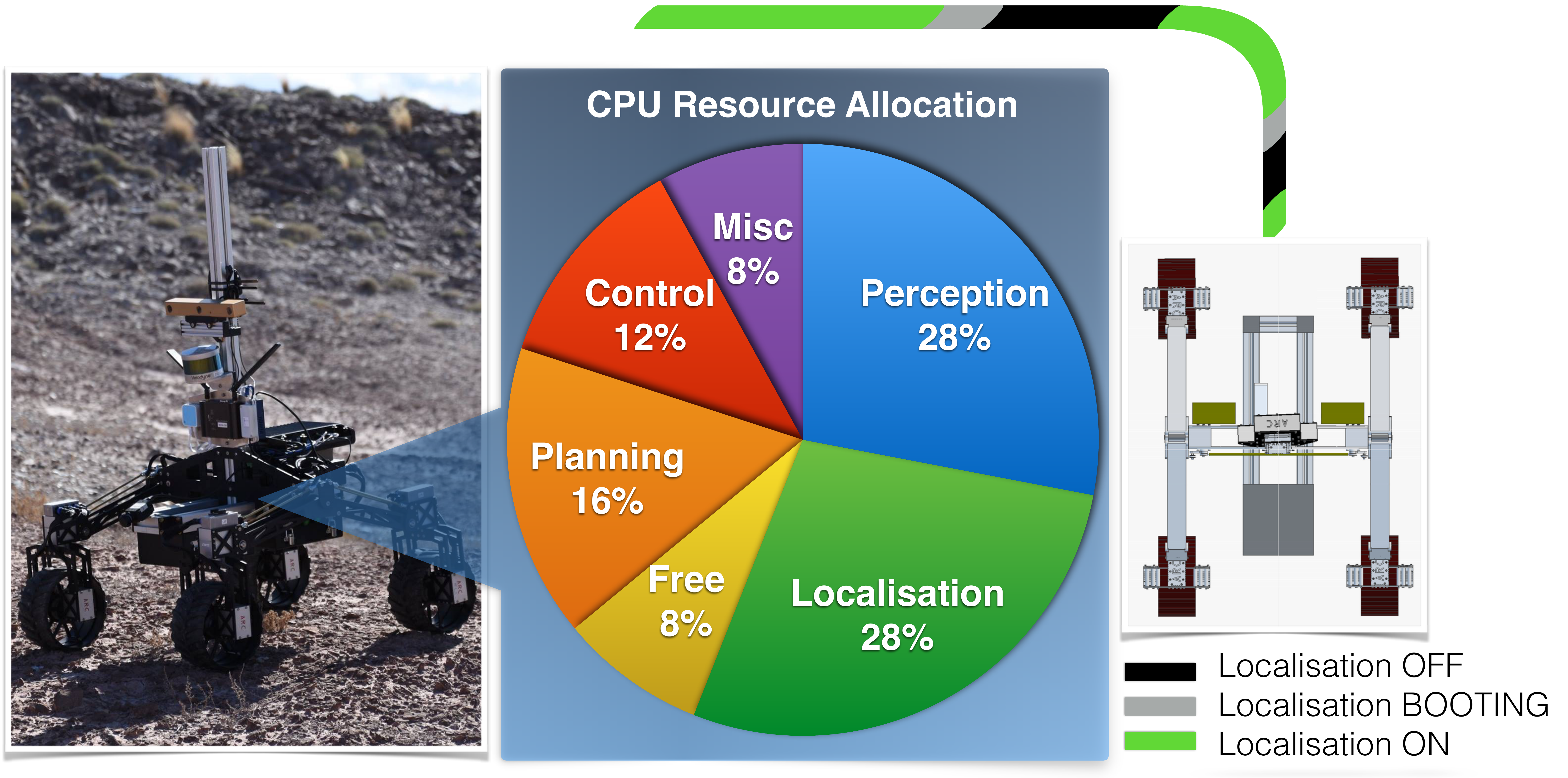}}
% \vspace{-5 mm}
\caption{
\footnotesize
A robot with a stereo camera with an on-board processor running various modules required for mobile autonomy.
The color-coded pie chart is an example representation of per-module CPU usage. By intelligently scheduling each module, we are likely to free resources, not only to reduce the resource budget, but also to free resources for other modules.
}
\vspace{-5 mm}
\label{fig:Resources}
\end{figure}

Mobile autonomy is enabled through {\em localization}, {\em perception} and {\em planning} modules, where localization and perception provide information about the robot's location and surroundings, respectively, and the planning module generates a trajectory. Most mobile robots treat these modules as separate processes, which are run simultaneously, and often continuously, for best performance.  These algorithms are complex and consume computational resources in addition to the energy cost of the robot's motion (motors). An example of CPU usage by the modules for a mobile ground robot is shown in Fig. \ref{fig:Resources}.  By intelligently scheduling the modules \cite{ondruska_icra15}, it may be possible to reduce the resource consumption while maintaining best performance.  More interestingly, it may be possible to \emph{trade off} reduced resource consumption for a slightly lower but still acceptable level of performance. Examples include switching localization on and off to save energy or restricting the continuous calls of the planner to free resources for other modules. One issue with this approach is that the objectives, \ie to reduce resource usage and to improve performance, are naturally competing, and by optimizing for one objective, the values for the other may become suboptimal. For instance, by turning localization off throughout the trajectory, the robot may minimize its energy consumption, while at the same time increase the probability of collision. On the other hand, keeping localization on for the duration can lead to excessive energy usage.

The aim of this paper is to provide a framework to aid the designers in exploring such \emph{resource-performance trade-offs} and finding schedules for mobile robots, guided by questions such as ``given a resource budget, what guarantees can be provided on achievable performance?'' and, more interestingly, ``what is the minimum resource budget required to achieve a given level of performance?''. To this end, we exploit a technique from formal methods known as quantitative \emph{multi-objective} verification and controller synthesis \cite{multiomdps_tacas07,multiomdps_atva12}, which, for a given scenario and a set of quantitative objectives, \eg constraints on computation power, time, energy, or probability of collisions, produces the so-called \emph{Pareto front}, a set of Pareto optimal points that represents all the optimal trade-offs that are achievable. The designer then selects a specific Pareto point based on the resource constraints and desired performance level, and a correct-by-construction schedule that meets those constraints is \emph{automatically} generated.  
We illustrate the potential of the framework on a generic resource cost function in conjunction with two performance criteria of safety and target-reachability in the context of a localization schedule but emphasize that the technique is generally applicable.
% We illustrate the potential of the framework on an example focused on energy optimization in the context of a localization schedule with three objectives but emphasize that the technique is generally applicable.

Multi-objective techniques have been studied for probabilistic models endowed with costs such as energy or time.  Off-the-shelf tools exist to support Pareto front computation \cite{prism,prism-games}, but are limited to discrete models such as Markov decision processes (MDP). The challenge here, therefore, is to adapt the techniques to the \emph{continuous-domain}, both time and space, scenario typical for mobile robots, while also capturing the uncertainty that the autonomy modules must deal with. We thus propose a novel \emph{abstraction} method, which considers noise in both the robot dynamics and its sensors.  Given a set of control laws, the method obtains a \emph{discretization} of the continuous robot motion as an MDP. We lift the robot's resource consumption to a cost on this MDP, and through the multi-objective techniques, we compute the Pareto front that encodes all optimal trade-offs between the resource requirements and the task-related performance guarantees.  Having this set available to the designer, they have the freedom to choose a Pareto point based on their design criteria.  For the selected point, we generate a schedule.  We demonstrate the efficacy of this framework on several robotic scenarios with various dynamics and resources.  
%; the accuracy of our abstraction is within 5\% of the theoretical result. Add how the designer chooses the point?
Summarizing, the main contributions of this work are:
\begin{itemize}
\item a general framework for the exploration of resource-performance trade-offs in mobile robotics based on multi-objective optimization with various (possibly conflicting) objectives,
\item a discretization of the robot dynamics that enables reduction to the multi-objective problem, and
\item validation of the techniques by both simulation and experiments of complex robotic scenarios with encouraging results.
\end{itemize}

\section{Literature Review} \label{lit_rev}
Most existing resource allocation works in robotics focus on single objective problems, namely optimization for QoS or energy.  Examples include \cite{MadanTON2010} in wireless systems, \cite{NerurkarICRA09} in multi-robot localization, and \cite{ondruska_icra15} in perception.  Such problems usually involve scheduling sensor usage, and the typical performance criterion is the ``goodness'' of the state estimation, \eg \cite{MR06,GCHM06}.  Apart from \cite{cloud-tradeoff}, where computation is offloaded to the cloud to improve performance, trade-off analysis between resource and performance objectives has been little studied.  Our framework not only performs such analysis, but it also allows multiple objectives for various resources, \eg energy, time, and computation power, and performance criteria, \eg safety and target reachability.

% To the best of our knowledge, computational resource allocation in the domain of mobile robotics is largely an unexplored area. Prior work considered specific resource allocation problems (QoS or energy), in wireless systems \cite{MadanTON2010}, multi-robot localization \cite{NerurkarICRA09} and perception \cite{ondruska_icra15}. Apart from \cite{cloud-tradeoff}, where computation is offloaded to the cloud to improve performance, trade-off analysis has been little studied. \rev{}{Specifically in sensor scheduling the resource to be allocated is the measurement frequency or other sensor usage cost. Approaches exist to generate optimal sensor frequencies~\cite{MR06} or sensor usage schedules~\cite{GCHM06} in order to optimize state estimation, but these do not allow goals such as safety or their combination and do not offer trade-off analysis.}

Since mobile robots are increasingly often employed in safety- and performance-critical situations, techniques from formal methods that offer high-level temporal logic planning \cite{Hadas:ICRA:2007}, correct-by-construction controller synthesis \cite{Wongpiromsarn:HSCC:2010} and performance guarantees \cite{Lahijanian:ICRA:2009,Luna:WAFR:2014} have been gaining attention. However, these are task specific and lack the generality and flexibility of the proposed framework, which enables systematic exploration of trade-offs.  

Trade-off analysis techniques have been studied in verification, mostly from the theoretical perspective, and include quantiles and multi-objective methods. Quantiles can express the cost-utility ratio but not the Pareto front, and have been applied for energy analysis of low-level protocols \cite{Baier2014}. Multi-objective (or multi-criteria) optimization has been extensively studied in operations research and stochastic control \cite{climaco}. More recently, techniques that combine high-level temporal logic specifications with multi-objective optimization have been formulated for discrete probabilistic models, including probabilistic \cite{multiomdps_tacas07} and expected total reward \cite{FKN+11,multiomdps_atva12} properties used in this paper. They have been employed, e.g., to analyze human-in-the-loop problems \cite{Feng-et-al}. %, but not in robotics. 
The works \cite{Chatterjee:HSCC:2008,Tesarova:ACC:2016} consider problems with budget constraints for discrete models from the theoretical point of view.

% \todo{Add a paragraph on POMDPS: FIRM, rate optimization, etc.}

\section{Problem formulation}\label{sec:pf}
The focus of this work is an optimal trade-off analysis between a robot's resource usage and its task guarantees through the use of a localization module. We consider robot dynamics (plant model) given by:
% Consider a mobile robot that moves in an environment according to the following equation of motion (plant model):
\begin{equation}\label{eq:dynamics}
	\dot{\vx} = f(\vx,\vu,\vw),
\end{equation}
where $\vx \in X \subseteq \real^{n_x}$ is the state of the robotic system, $\mathbf{u} \in U \subseteq \real^{n_u}$ is the control input, $\vw \in \real^{n_w}$ is the process (motion or input) noise given by a normal distribution $\N(\mathbf{0}, \mQ)$ with zero mean and covariance $\mQ \in \real^{n_w \times n_w}$, and $f:X \times U \times \real^{n_w} \rightarrow \real^{n_x}$ is a continuous integrable function that describes the evolution of the robot in the space.   The robot is equipped with two sets of sensors that enable the measurement of its state, \eg odometry and high-accuracy localization sensors.  The first set (odometry) uses a negligible amount of resources but provides inaccurate noisy measurements. By deploying the second set, which we refer to as the localization module or simply \emph{localization}, additional information is obtained, and the measurements become more accurate at a higher cost of resources.  

% Let $A=\{a_\start,a_\boot,a_\on,a_\off\}$ denote the set of all possible localization module actions (status). If the module is initially off, it takes some time for it to boot up. Action $a_\start$ sends a signal to start the localization, and it takes time $T_\boot\in \real_{\geq 0}$ for it to turn on.  The action $a_\boot$ refers to the booting status of localization during $T_\boot$.  Action $a_\on$ indicates that localization is on, resulting in precise measurement of the current state of the system. Finally, $a_\off$ turns the localization off. The resulting measurement model is:
Let $A=\{a_\start,a_\boot,a_\on,a_\off\}$ denote the set of all possible localization module actions (status).  Action $a_\start$ sends a signal to start the localization, and it takes time $T_\boot\in \real_{\geq 0}$ for it to turn on.  The action $a_\boot$ refers to the booting status during $T_\boot$.  Action $a_\on$ indicates that localization is on, and $a_\off$ turns it off. The resulting measurement model is:
% \begin{equation} \label{eq:sensor}
% 	\vz = h(\vx,\vv),
% \end{equation}
% where $\vz \in \real^n$ is the state measurement, $h:X \times \real^n \rightarrow \real^n$ is the measurement process function, and $\vv$ %:A\to \real^n$ 
% is a random variable with values in $\real^n$ representing sensors' noise as%given by
% \begin{equation}\label{eq:noise}
% 	\vv = 
% 	\begin{cases}
% 		\mathbf{0} & \text{if }a = a_\on, \\ 
% 		\text{a sample from }\V & \text{if }a \in \{a_\start,a_\boot,a_\off \},
% 	\end{cases} 
% \end{equation}
% where $\V$ denotes the distribution of the noise of the first set of sensors (odometry). 
\begin{equation}
	\label{eq:sensor}
	\vz=\begin{cases}
		h^\od(\vx,\vv^\od) & \text{if }a \in \{a_\start,a_\boot,a_\off \}, \\
		h^\lo(\vx,\vv^\lo) & \text{if }a = a_\on,
	\end{cases}
\end{equation}
where $\vz \in Z \subseteq \real^{n_z}$ is the state measurement, and $\vv^\od, \, \vv^\lo \in \real^{n_v}$ are the measurement noise terms under the first and second set of sensors (odometry and localization), respectively.  In high accuracy mode, the noise is given by $\vv^\lo \sim \N(\mathbf{0},\mR)$, where covariance $\mR \in \real^{n_v \times n_v}$, whereas in low accuracy mode, no restriction is imposed on the distribution of $\vv^\od$.

% For the method herein, we assume that 
The robot moves in an environment (workspace) $W \subset \real^{n_W}$, where $n_W \in \{2,3\}$, with a set of obstacles $W_O$ and a target region $W_G$.  Colliding with an obstacle constitutes failure; hence, the robot's task is to avoid obstacles and reach the target by following a precomputed reference trajectory $\traj$. We assume that $\traj$ is given as a sequence of waypoints $\traj= (\tilde{\vx}_1, \ldots, \tilde{\vx}_{|\traj|})$, where $\tilde{\vx}_i \in X$. We denote the initial state of the robot by $\tilde{\vx}_0$.  The robot is equipped with an on-board controller that generates a sequence of control laws in order to follow $\traj$ (see Sec. \ref{sec:feedback}).  

% We assume that, when the measurements $\vz$ are precise, \ie $a=a_\on$, the $i$-th control law drives the robot to waypoint $\tilde{\vx}_{i}$.  When the robot turns its localization module off to conserve energy, it may deviate from $\traj$, \ie the robot may not end exactly on $\tilde{\vx}_i$ after the execution of the $i$-th control law due to noisy measurements.  Therefore, even if $\traj$ is an obstacle-free trajectory, by turning the localization off, there is a probability that the robot may collide with an obstacle or may not reach the target. We specify these probabilities formally in Sec.~\ref{sec:performance}.  
We assume that, when the localization module is used, \ie $a=a_\on$, the $i$-th control law drives the robot to a proximity of waypoint $\tilde{\vx}_{i}$ (see Sec. \ref{sec:feedback}).  When the robot turns its localization off, it may deviate from $\traj$ due to imprecise measurements.  Therefore, even if $\traj$ is an obstacle-free trajectory, by turning off the localization module, there is a probability that the robot may collide with an obstacle or may not reach the target. We specify these probabilities formally in Sec.~\ref{sec:performance}.

The aim is, given $\traj$, to schedule the use of localization over time in a way that optimizes the trade-off between the robot's resource usage and its task guarantees.
% The goal is, given $\traj$, to compute all the optimal trade-offs between the robot's resource usage and its task guarantees through localization scheduling.  
Let $\sched$ denote a localization schedule, which simply speaking, indicates which localization action the robot needs to apply at any given time (see Sec.~\ref{sec:locsched}).  The resource consumption of the robot under schedule $\sched$ is the sum of the resources required for it to run the localization module and the resources consumed by the rest of the system. 
% While different types of resources can be considered, such as computational power or energy source, we consider the the total energy consumption as a unifying metric and argue that it can be utilized to encompass both as detailed in Sec.~\ref{sec:performance}. 
To allow the inclusion of different types of resources, \eg computational power, time, or energy, we define a general cost function $c: X \times U \times A \rightarrow \real_{\geq 0}$ (see Sec.~\ref{sec:energy} for details and Sec.~\ref{sec:cs} for examples) to represent total resource usage.  The formal problem definition is then as follows.

% Without loss of generality, we attribute the latter mainly to the motion, \ie motors. Therefore, the expected total energy under schedule $\sched$ is
% \begin{equation}\label{eq:E}
% 	E_T(\sched) = E_S(\sched) + E_M(\sched),
% \end{equation} 
% where $E_S$ and $E_M$ denote the expected energy consumption by the sensors and motors, respectively.  
% We expand on the details of localization schedule and total energy in Sec. \ref{sec:locsched} and~\ref{sec:energy}, respectively.  

% The formal definition of the problem follows.
\begin{problem} \label{probform}
% 	Given a mobile robot with the equation of motion in the form of \eqref{eq:dynamics} and measurement model in the form of \eqref{eq:sensor} in an environment $W$ with a set of obstacles $W_O$ and a target $W_G$, and a reference trajectory $\traj$ with its corresponding control laws, compute a localization schedule $\sched$ such that
    Given a mobile robot model as in \eqref{eq:dynamics} and \eqref{eq:sensor} in an environment $W$ with a set of obstacles $W_O$ and a target $W_G$, a reference trajectory $\traj$ with its corresponding control laws, and resource cost function $c$, compute a localization schedule $\sched$ such that
	\begin{itemize}
		\item the expected total resource cost is minimized,
		\item the probability of collision is minimized, and
		\item the probability of reaching the target is maximized.
	\end{itemize}
\end{problem}

\smallskip

% It should be noted that the objectives in Problem \ref{probform} may be competing, \ie by optimizing for one objective the values of the other two may become suboptimal. For example, turning off the localization module may decrease the energy consumption but may also result in a poor performance in terms of avoiding collision and reaching the target.  On the other hand, by using perfect localization, the task performance may become optimal, but the total energy consumption may be unnecessarily high. Therefore, there may not exist a localization schedule that globally optimizes all three objectives.  
% In this work, we propose a framework to compute the set of all optimal trade-offs and their corresponding schedules. Then, the designer can choose the schedule that gives rise to their preferred optimal trade-off of the objectives.  

These objectives may be competing, and there may not exist a localization schedule that globally optimizes all of them.
In this work, we are interested in the set of all optimal trade-offs between the objectives, which introduces an additional level of complexity to the problem.  Another major challenge is dealing with a continuous robotic system with noisy measurements, \ie partial observability.  This leads to reasoning in belief space, which is generally a computationally infeasible domain.  We propose a framework to address these challenges in two steps.  First, we overcome the complexity of belief space through a suitable finite abstraction and then use formal techniques to generate the set of all optimal trade-offs between the  objectives.  

% This framework is general in that its structure is independent of the choices of robot model and objectives.  It can also handle multiple resource cost functions. Here, we present the concrete design of the two steps of the framework for the particular choices in Problem~\ref{probform} but note that, in addition to more objectives, it can also be adapted to schedule other modules such as perception or different motors.
This framework is general in that its structure is independent of the choices of objectives.  It can also handle multiple resource cost functions. Here, we present the concrete design of the two steps of the framework for the particular choices in Problem~\ref{probform} but note that, in addition to more objectives, it can also be adapted to schedule other modules such as perception or different motors.

\section{System Description}\label{sec:sysdes}
% In this section, we provide a detailed description of the system and formal definitions for the notions used above.

%The robot moves in a static environment (workspace) $W \subset \real^{n_W}$, where $n_W \in \{2,3\}$, with a known map.  The workspace consists of a set of convex obstacles $W_O = \{W_{o_1},\ldots,W_{o_{|W_O|}}\}$, a goal set $W_G$, and the free space $W_F = W - (W_O \cup W_G)$.  To establish the relationship between the state space $X$ of the robotic system and its workspace $W$, let the first $n_W$ components of state $\vx \in X$ refer to the position of the robot.  Then, $W$ is the projection of $X$ onto $\real^{n_W}$.  We denote the function that performs this projection by $\proj: X \rightarrow W$.  Similarly, given $\vw \in W$, we define $\proj^{-1}(\vw) = \{ \vx \in X \mid \proj(\vx) = \vw \}$.  Therefore, the three sets of $W_O$, $W_G$, and $W_F$ induce a partition in the state space denoted by $X_O=\{X_{o_1},\ldots,X_{o_{|W_O|}}\}$, $X_G$, and $X_F$, respectively.  Having established this relationship, we focus on $X$ for the remainder of the paper.

Due to both process and measurement noise, robot's motion is stochastic, and its exact state cannot be known.  They, however, can be described as probability distributions. The probability distribution of $\vx_t$ at time $t \in \real_{\geq 0}$ is referred to as the \textit{belief} of $\vx_t$, denoted by $\Bz_t$, and given by 
\begin{equation}
\label{eq:belief}
	\vx_t \sim \Bz_t = \pdist(\vx_t \mid \vx_{t_0}, \, \vu_{t_0:t}, \, \vz_{t_0:t}, \, \loc_{t_0:t}),
\end{equation}
where $\pdist$ denotes the (conditional) probability density function of $\vx$, $\vx_{t_0}$ is the distribution at the initial time $t_0$, and $\vu_{t_0:t}$, $\vz_{t_0:t}$, and $\loc_{t_0:t}$ are the sequences of control inputs, measurements, and statuses of the localization used from $t_0$ to $t$. 
We denote the belief space containing all possible beliefs by $\Bspace$, \ie $\Bz_t \in \Bspace$ $\forall t$. 

\subsection{Control Laws}
\label{sec:feedback}
Recall that, starting from the initial position $\tilde{\vx}_0$, the robot follows reference trajectory $\traj = (\tilde{\vx}_1, \ldots, \tilde{\vx}_{|\traj|})$ using a series of \textit{control laws}.  
% This series is constructed by associating each waypoint $\tilde{\vx}_i$, with $1 \leq i \leq |\traj|$, with a \textit{control law} (also known as \textit{control symbol}) $\tilde{\vu}_i = (g_i,\trigger_i)$,
% \begin{equation*}
% 	\tilde{\vu}_i = (g_i,\trigger_i),
% \end{equation*}
For $1 \leq i \leq |\traj|$, control law  $\tilde{\vu}_i = (g_i,\trigger_i)$ consists of a feedback controller  $g_i: \Bspace \rightarrow U$  
designed to drive the robot towards $\tilde{\vx}_i$
and a termination rule (trigger) $\trigger_i$ that indicates when to terminate the execution of $g_i$.  
We assume that, when localization is on, $g_i$ is able to stabilize the state belief $\Bz$ around waypoint $\tilde{\vx}_i$.  The construction of such a controller for robotic systems is detailed in \cite{Agha:IJJR:2014}.  In short, $g_i$ is generally a concatenation of two controllers: reachability and stabilizer.  The reachability controller drives the system to a neighborhood of $\tilde{\vx}_i$, and then the stabilizer controller stabilizes $\Bz$ to a predefined belief $\tilde{\B}_i$ that corresponds to $\tilde{\vx}_i$.  This stabilization is typically achieved by an LQG controller on the linearized dynamics around $\tilde{\vx}_i$ and defining $\tilde{\B}_i = \N(\tilde{\vx}_i,\cov_{\tilde{\vx}_i})$, where $\cov_{\tilde{\vx}_i}$ is the steady-state covariance given by steady-state Kalman filter (solution to an algebraic Riccati equation \cite{Bertsekas:DP:2007}).  Note that the convergence to $\tilde{\B}_i$ is guaranteed if the linearized dynamics are controllable and observable \cite{Bertsekas:DP:2007}.  For a full discussion on the construction of such controllers for various systems, including nonholonomic systems, see \cite{Agha:IJJR:2014}.  When localization is off, $g_i$ uses only the reachability part of the controller.  

% Let $t_{i-1}$ and $\Delta t_i$ be the time mark of the initialization of $\tilde{\vu}_i$ and the duration that it takes for the robot to move from $\tilde{\vx}_{i-1}$ to $\tilde{\vx}_i$ under $g_i$ with localization on, respectively.  
Let $\Delta t_i$ be the duration that it takes the reachability controller to move the robot from $\tilde{\vx}_{i-1}$ to (a neighborhood of) $\tilde{\vx}_i$ under localization on.  We design the trigger $\trigger_i$ to fire, \ie $\trigger_i=1$, when the belief of the robot state converges to $\tilde{\B}_{i}$ if the localization is on (in practice, an $\epsilon$-convergence is sufficient, \ie $|\Bz_t - \tilde{\B}_{i}| < \veps$).  
In turn, if the localization is not on, the robot applies $g_i$ for the duration of $\Delta t_i$. % and then switches to the next control law. We define the termination rule to be
Formally, 
\begin{equation}\label{eq:termrule}
	\trigger_i =
	\begin{cases}
		\mathbbm{1}_{<\veps}(|\Bz_t - \tilde{\B}_{i}|) 	& \text{if }a=a_\on, \\
		\delta(t - (t_{i-1} + \Delta t_i))  & \text{if }a \in \{a_\start,a_\boot,a_\off \},
	\end{cases}
\end{equation}
where $\mathbbm{1}$ and $\delta$ are the Indicator and Dirac delta functions, respectively, and $t_{i-1}$ is the time mark of the initialization of $\tilde{\vu}_i$. 
Furthermore, $\veps = (\epsilon_\mean, \epsilon_\var)^T \in \real^2_{> 0}$, and for $\Bz_t = \N(\hat{\vx}_t, \cov_{\tilde{\vx}_i})$, where $\hat{\vx}_t$ and $\cov_{\tilde{\vx}_i}$ are the state estimate and covariance at time $t$, the indicator function $\mathbbm{1}_{<\veps}(|\Bz_t - \tilde{\B}_{i}|) = 1$ if $|\hat{\vx}_t - \tilde{\vx}_i| < \epsilon_\mean$ and $|\cov_{x_t} - \cov_{\tilde{\vx}_i}| < \epsilon_\var$; otherwise 0.
%Under control law $\tilde{\vu}_i$, the robot executes controller $g_i$ until $\trigger_i$ fires, \ie $\trigger_i=1$.  With this design of $\trigger_i$, $\tilde{\vu}_i$ guarantees that, when the localization is on, the robot ends exactly on $\tilde{\vx}_{i}$.  In turn, when the localization is not on, the robot applies $g_i$ for the duration of $\Delta t_i$ and then switches to the next control law.  
% We note that the introduce  this without loss of generality as in reality a decision still needs to be taken as to whether a waypoint has been reached or not. In fact, it is sufficient to reach a proximity of the waypoint, as discussed in Sec.~\ref{sec:concl}. We simply use this assumption, for simplicity of presentation.
Therefore, from trajectory $\traj$, the following series of control laws, which enable the robot to follow $\traj$, is obtained:
\begin{equation} \label{eq:trajcontrol}
	\traj= (\tilde{\vx}_1, \ldots, \tilde{\vx}_{|\traj|}) \quad \Rightarrow \quad \vu = (\tilde{\vu}_1, \ldots, \tilde{\vu}_{|\traj|}).
\end{equation}

% \todo{
% Separate controllers for $a=a_\on$ and $a \in \{a_\start,a_\boot,a_\off \}$:
% \begin{enumerate}
% \item $a=a_\on$: FIRM framework, introduce briefly. For every waypoint $\tilde{\vx}_i$, a controller $\tilde{\vu}^\lo_i$ (function + trigger) is given that drives the system from any initial belief to a concrete belief $b_i$ (distance of the mean and covariance is the trigger) around waypoint $\tilde{\vx}_i$. Belief at time $t$ given an initial state $\vx_{t_0}$, and a history of localization actions $a_{t_0:t}$ and observations $\vz_{t_0:t}$ is defined as follows:
% \begin{equation}\label{eq:bbelief}
% \vx_t\sim b_t=\pdist(\vx\mid  \vx_{t_0}, \, a_{t_0:t}, \, \vz_{t_0:t}).
% \end{equation}
% \item $a \in \{a_\start,a_\boot,a_\off \}$: no strong assumptions on controllers, $\tilde{\vu}^\od_i$ is a function ($\distr(X)\to U$) and a trigger (could be time or distance, etc) 
% \end{enumerate}
% }

%%%%%%%%%%%%%%%%%%%%%%%%%%%%%%%%%%%%%%%%%%%%%%%%%%%%%%%%%%%%%%%%%%%
%%%%%%%%%%%%%%%%%%%%%%%%%%%%%%%%%%%%%%%%%%%%%%%%%%%%%%%%%%%%%%%%%%%

\subsection{Localization Schedule}
\label{sec:locsched}

% Robot's motion becomes stochastic under control law $\tilde{\vu}_i$ when the localization is not on, and its exact state cannot be known due to sensor noise.  The robot state, however, can be described as a probability distribution. This probability distribution of $\vx_t$ at time $t \in \real_{\geq 0}$ is referred to as the \textit{belief} of $\vx_t$, denoted by $\B_t$, and given by 
% \begin{equation}
% \label{eq:belief}
% 	\vx_t \sim \B_t = \pdist(\vx_t \mid \vx_{t_0}, \, \vu_{t_0:t}, \, \loc_{t_0:t}),
% \end{equation}
% \todo{don't define here, only in solution} 
% where $\pdist$ denotes the (conditional) probability density function of $\vx$, $\vx_{t_0}$ is the distribution at the initial time $t_0$, and $\vu_{t_0:t}$ and $\loc_{t_0:t}$ are the sequences of control inputs and statuses of the localization used from $t_0$ to $t$. Therefore, the robot has to make a decision on the use of its localization based on $\B_t$.  This decision is referred to as the \textit{localization schedule}. Let $\distr(\cdot)$ denote the set of all probability distributions over a given set.  Localization schedule is a function $\sched : \distr(X) \to \distr(A)$ that assigns to a belief $\B_{t}$ a probability distribution over the localization actions from the set $A$.

The robot makes a decision on the use of its localization based on its belief $\Bz_t$.
This decision is referred to as the \textit{localization schedule}. Let $\distr(\cdot)$ denote the set of all probability distributions over a given set.  Localization schedule is a function $\sched : \Bspace \to \distr(A)$ that assigns to a belief $\Bz_{t}$ a probability distribution over the localization actions from the set $A$. 
% \todo{localization schedule is a function $\sched : \distr(X) \to \distr(A)$ that assigns to a belief $b_{t}$, see Eq.~\ref{eq:bbelief}, a probability distribution over the localization actions from the set $A$}

In this work, we assume that the localization decisions are made right before applying each control law $\tilde{\vu}_i$. In other words, the granularity of the localization decisions corresponds to the waypoints in $\traj$, which is user-defined.
% \todo{Move either to problem formulation or after solution: discussion on the granularity of the trajectory and the connection to the original problem formulation.}
Therefore, given a reference trajectory $\traj$ and a localization schedule $\sched$, the induced robot trajectory can be described by a sequence of state beliefs:
\begin{eqnarray}
\label{eq:beltraj}
	% \Bz_{t_0} \xrightarrow{(\loc_{t_0:t_1},\tilde{\vu}_1)} \Bz_{t_1} \xrightarrow{(\loc_{t_1:t_2},\tilde{\vu}_2)} \ldots \hspace{30mm} \nonumber \\
	% \xrightarrow{(\loc_{t_{|\traj|-1}:t_{|\traj|}},\tilde{\vu}_{|\traj|})} \Bz_{t_{|\traj|}},
	\Bz_{t_0} \xrightarrow{(\loc_{t_0:t_1},\tilde{\vu}_1)} \Bz_{t_1} \xrightarrow{(\loc_{t_1:t_2},\tilde{\vu}_2)} \ldots 
	\xrightarrow{(\loc_{t_{|\traj|-1}:t_{|\traj|}},\tilde{\vu}_{|\traj|})} \Bz_{t_{|\traj|}},
\end{eqnarray}
where the localization action of $a_{t_i}$ is assigned according to the probability distribution $\sched(\Bz_{t_i})$ over $A$.  
% Intuitively, a schedule decides on the use of localization sensors between time points $t_i$ and $t_{i+1}$ based on the belief $\B_{t_i}$. \maja{[remove the last sentence? sounds repetitive]}
Note that it is possible for the localization to become active during the execution of $\tilde{\vu}_i$ if $a_{t_{i-1}} = a_\boot$. That means that, at some point between $t_{i-1}$ and $t_i$, booting is complete.
% At this point, the exact state of the robot becomes known, and $\tilde{\vu}_i$ drives the robot to $\tilde{\vx}_i$.
 % and hence $\B_{t_i}(\vx_{t_i} = \tilde{\vx}_i)$ = 1. \maja{[use $\B_{t_i}=\delta(\vx_{t_i}-\tilde{\vx}_i)$ instead?]}  
At this point, the measurements become more accurate, and $\tilde{\vu}_i$ drives the robot's state belief to $\tilde{\B}_i$.

We assume that, without loss of generality, the localization is initially on.  Therefore, in Problem~\ref{probform}, we are interested in computing a \emph{feasible} schedule, in which it holds that: 
%\begin{itemize}
the first action applied by the schedule is $a_\on$ or $a_\off$; % \ie $\sched(\B_{t_0})(a_\start) = \sched(\B_{t_0})(a_\boot) = 0$,
every action $a_\start$ is immediately preceded by $a_\off$; 
every action $a_\boot$ is immediately preceded by $a_\start$ or $a_\boot$; 
if $a_\start$ is used at time point $t_i, i\geq 1$, then action $a_\boot$ is used at all time points $t_{i+1}, \ldots ,t_{j}$ such that $t_{j}-t_i< T_\boot \leq t_{j+1}-t_i$; 
and every action $a_\on$ is immediately preceded by $a_\boot$ or $a_\on$. 
%\end{itemize} 

Since we are interested in efficient schedules, we assume that all schedules avoid turning off the localization module while booting is in progress and starting the booting process if it cannot be completed by the time $t_{|\traj|}$. 

\subsection{Objectives}\label{sec:performance}

\subsubsection{Resource Consumption}
\label{sec:energy}
One of the objectives that influences the choice of $\sched$ is the resource consumption.  Let $c: X \times U \times A \rightarrow \real_{\geq 0}$ denote a resource consumption function, which represents the amount of resources used by the robotic system given its state, control input, and localization action.  An example of such resource cost is the amount of computations required by different system modules (including localization).   We are interested in the total expected resource cost under $\sched$, which we denote by $E_\cost(\sched)$.  

Let $\B_t$ denote the \textit{expected belief}, where expectation is taken over observations, \ie 
\begin{align}
	\B_t &= \expv_Z(\Bz_t\mid \vx_{t_0}, \, \loc_{t_0:t}) \\
	&= \int_{Z_{t_0:t}} \pdist(\vx_t \mid \vx_{t_0}, \loc_{t_0:t}, \vz_{t_0:t}) pr(\vz_{t_0:t}) d\vz_{t_0:t}\\
	&= \pdist(\vx_t \mid \vx_{t_0}, \, \loc_{t_0:t}),
\end{align}
where $\expv_Z$ is the expectation over domain $Z$, and $pr$ is the probability measure.
Then, given the localization action $a_t$ and control $\vu_t$, the expected cost at time $t$ is given by
\begin{equation}\label{eq:expcostatt}
	\expv_X(c(\vx_t,\vu_t,a_t)) = \int_{X} c(\vx_t,\vu_t,a_t) \, \B_t \, d\vx_t,
\end{equation} 
where $\expv_X$ is the expectation over domain $X$.  
The total expected cost for the whole trajectory under $\sched$ is
\begin{equation} \label{eq:expcosttotal}
	E_\cost(\sched) = \int_{t_0}^{t_{|\traj|}} \sum_{a_t \in A} \sched(\B_t)(a_t) \, \expv_X(c(\vx_t,\vu_t,a_t)) \, dt.  
\end{equation}

\subsubsection{Task Performance}
\label{sec:coltar}
% The other objectives that influence the choice of $\sched$ are obstacle avoidance and reaching target.  
% These objectives need to be reasoned about probabilistically by using beliefs since the robot motion is generally stochastic under $\sched$.
%Recall that the robot trajectory is best described by its belief trajectory in \eqref{eq:beltraj}.  Therefore, even if the reference trajectory $\traj$ is obstacle-free, by turning off the localization process, there is a chance that the robot could collide with an obstacle. 
% The probability of collision at time $t$ is given by
% \begin{equation*}
% \label{eq:obsprobt}
% 	pr(\vx_t \in X_O) = \sum_{i=1}^{|W_O|} \int_{X_{o_i}} \B_t \, d\vx_t,
% \end{equation*}
% \maja{[remove sum, integral over $X_O$ enough. also, should not involve $\pdist$?]}
% where $X_O \subset X$ is a set of states that correspond to the obstacles in $W_O$, and $X_{o_i} \in X_O$ is a set of states corresponding to obstacle $i$, \ie $W_{o_i} \in W_O$.
% Given a schedule $\sched$, the probability of collision along trajectory $\traj$ is
% \begin{eqnarray}
% \label{eq:obsprob}
% 	P_\coll(\sched) &=& pr \big(\vx_{{t_0}:{t_{|\traj|}}} \in X_O \big) \nonumber \\
% 					&=& \int_{t_0}^{t_{|\traj|}} pr(\vx_t \in X_O) \, dt.
% \end{eqnarray}
The task objectives of obstacle avoidance and target reachability need to be reasoned about probabilistically by using beliefs.
Given that all collisions with obstacles are terminal, the fragments of the beliefs that are in collision remain in obstacles as $\B_t$ evolves.
Hence, the probability of collision along $\traj$ under schedule $\sched$ is
\begin{eqnarray}
\label{eq:obsprob}
    P_\coll(\sched) = pr \big(\vx_{{t_0}:{t_{|\traj|}}} \in X_O \big) = \int_{X_{O}} \B_{t_{|\traj|}} \, d\vx_{t_{|\traj|}},
% 	P_\coll(\sched) &=& pr \big(\vx_{{t_0}:{t_{|\traj|}}} \in X_O \big) \nonumber \\
% 					&=& \int_{X_{O}} \B_{|\traj|} \, d\vx_{t_{|\traj|}},
					% &=&  \sum_{i=1}^{|X_O|} \int_{X_{o_i}} \B_{|\traj|} \, d\vx_{t_{|\traj|}},
\end{eqnarray}
where $X_O \subset X$ is the set of states that correspond to the obstacles in $W_O$.
% , and $X_{o_i} \in X_O$ is a set of states corresponding to obstacle $i$, \ie $W_{o_i} \in W_O$.  
Similarly, the probability that the robot is in the target region at the end of the trajectory execution is
\begin{eqnarray}
\label{eq:tarprob}
    P_\targ(\sched) = pr(\vx_{t_{|\traj|}} \in X_G) = \int_{X_{G}} \B_{t_{|\traj|}} \, d\vx_{t_{|\traj|}},
% 		P_\targ(\sched) &=& pr(\vx_{t_{|\traj|}} \in X_G) \nonumber \\
% 						&=& \int_{X_{G}} \B_{t_{|\traj|}} \, d\vx_{t_{|\traj|}},
\end{eqnarray}
where $X_G \subset X$ is the set of states that correspond to the target region $W_G$.

\section{Solution Method}\label{sec:solution}
% To approach Problem~\ref{probform}, we first abstract the robot's execution to a finite probabilistic model through a novel discretization technique. We then use a reduction to a known problem of multi-objective optimization over the abstract model. Finally, leveraging recent results from formal verification, we construct the set of all solutions with optimal trade-offs to Problem~\ref{probform}.  %Finally, we employ the existing techniques and tools to solve the reduced optimization problem.

Here, we detail our abstraction method for the robotic system in Problem~\ref{probform} and the reduction of the problem to a multi-objective optimization over an MDP. 

\subsection{Abstraction}\label{subsec:abstraction}

\subsubsection{Belief Space Discretization} \label{sec:beliefDisc}

Recall that, due to motion and observation noise, the robot has to base its localization schedule on state beliefs.  These beliefs are generally hard to reason about because they are continuous in both time and space. One of the novelties of this work is a discretization method that reduces the complexity of this reasoning from continuous to a discrete, finite space.  The purpose is to capture all possible behaviors of the system in expectation; thus, we focus on the expected beliefs.  The key observation is that localization decisions are only made right before applying each control law, and there is only a finite number of control laws (waypoints) and localization actions. Hence, the number of belief states that the robot has to make a decision on is, in turn, finite. 
% In other words, the finiteness of the number of control laws and localization actions induces a discretization of the belief space with respect to the localization decision %time points.  

\begin{figure}[t]
\begin{center}
\subcaptionbox{Belief graph.\label{fig:beliefs}}{\scalebox{.36}{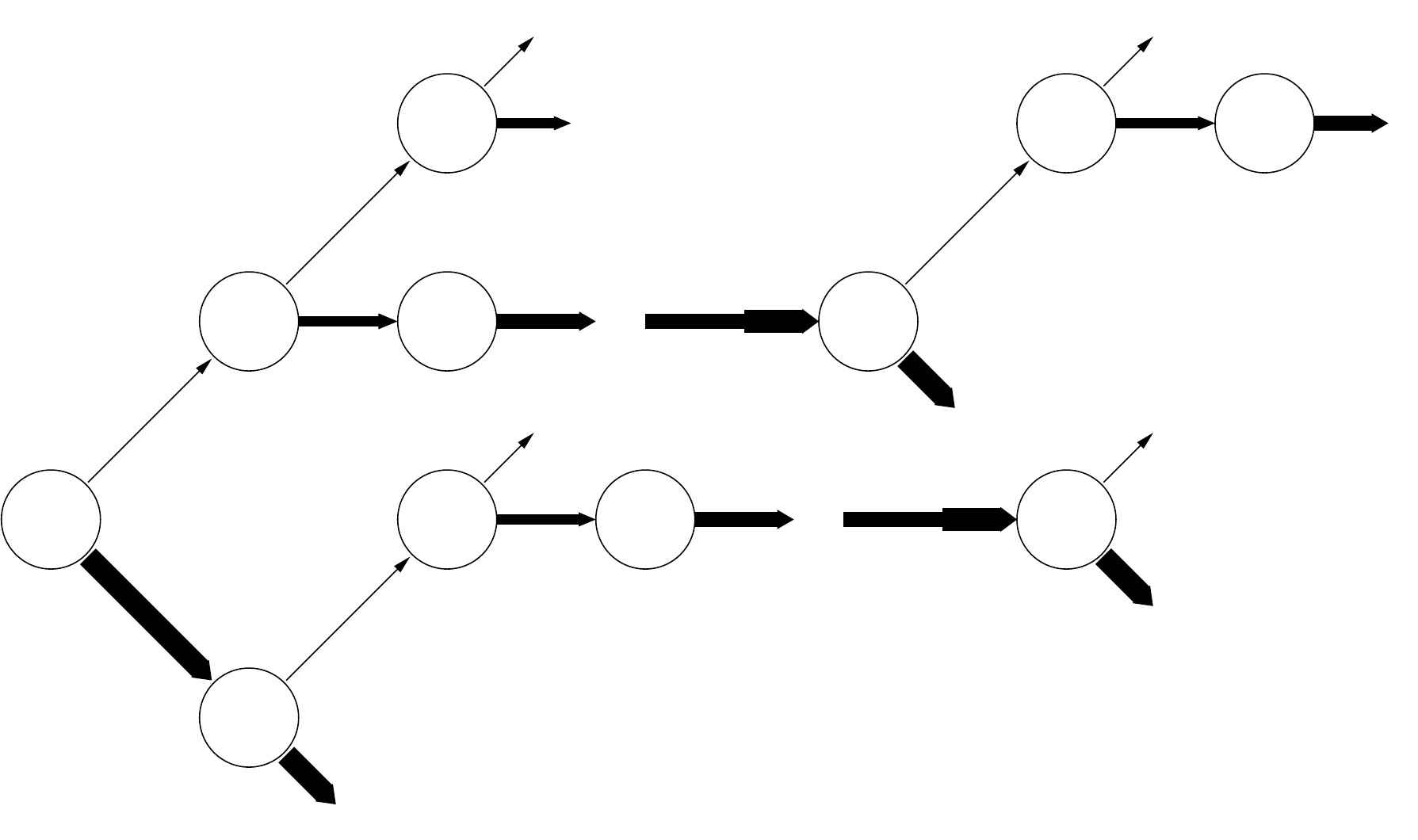}}
\subcaptionbox{Markov decision process.\label{fig:mdpstates}}{\scalebox{.43}{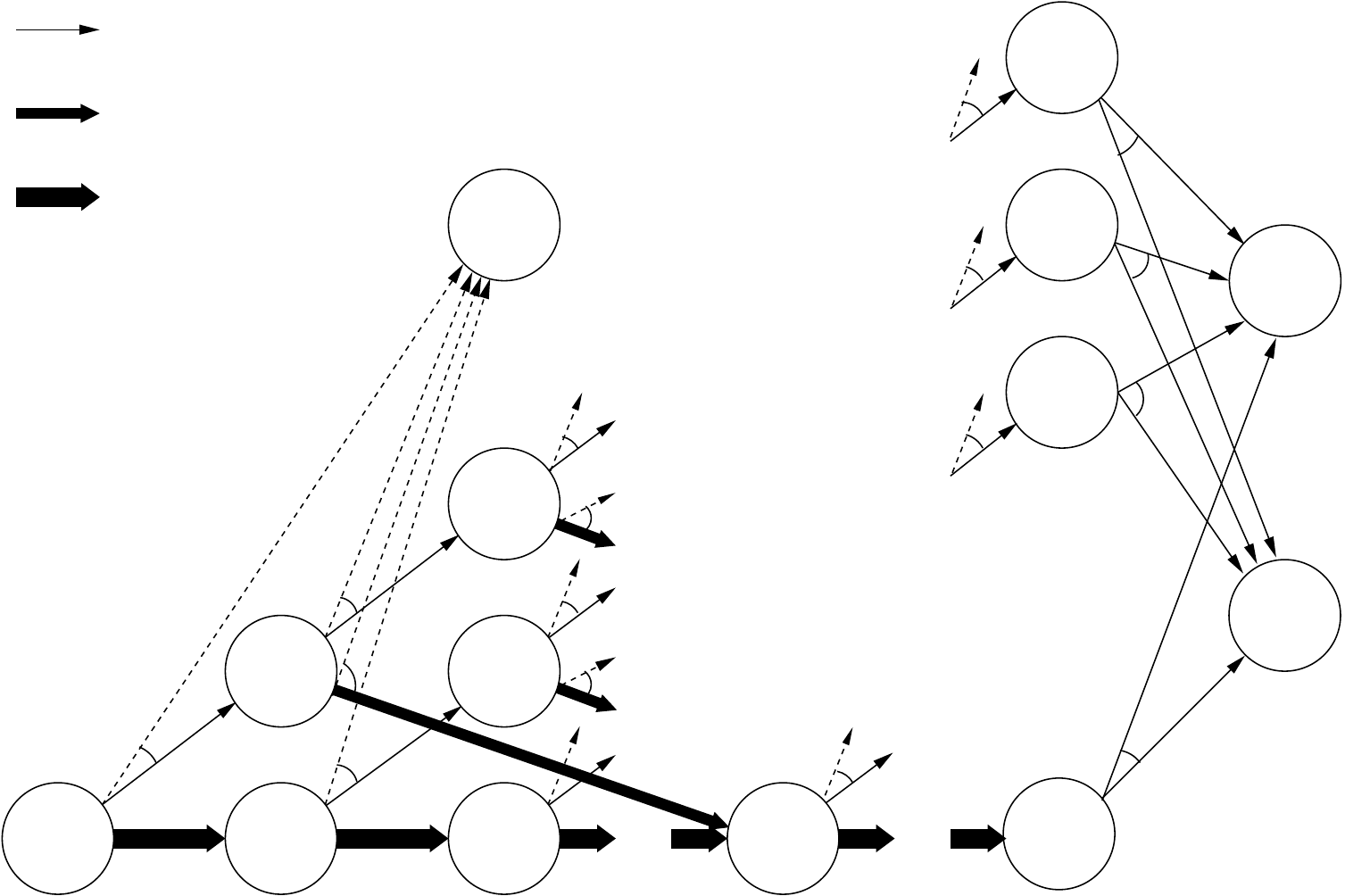}}
\end{center}
\caption{Structure of the belief graph and the MDP. The thickness of edges indicates their action label. 
% From thin to thick, these are $\loc_\off, \loc_\start, \loc_\boot, \loc_\on$ in the belief graph and $\loc_\off,\loc_\sbo$ and $\loc_\on$ in the MDP. 
In (b), action $\loc_\sbo$ represents the sequence of actions needed to fully activate localization, which begins with $\loc_\start$, continues with $\loc_\boot$, and ends with $\loc_\on$. 
% Transitions in a probabilistic distribution under a given action are shown with an arc, and dashed lines mark transitions to $s_\coll$. 
% Note that, in (b), we consider collision-free evolution of the robot under action $\loc_\on$.
}\label{fig:structure}
\end{figure}

% \todo{Introduce expected belief, where expectation is taken over observation:
% \begin{align*}
% \B_t &= E(b_t\mid \vx_{t_0}, \, \loc_{t_0:t})) \\
% &= \int_{Z_{t_0:t}} \pdist(\vx_t \mid \vx_{t_0}, \loc_{t_0:t}, \vz_{t_0:t}) p(\vz_{t_0:t}) d\vz_{t_0:t}\\
% &= \pdist(\vx_t \mid \vx_{t_0}, \, \loc_{t_0:t})
% \end{align*}
% These are the nodes of the belief graph.
% }
Technically, the robot's belief evolves sequentially based on the applied control law and localization action in each step as shown in \eqref{eq:beltraj}.  
%Since the localization schedule is called only at time point $t_{i-1}$ before applying control law $\tilde{\vu}_i$ for $i = 1,\cdots,|\traj|$, there are only a finite number of possible beliefs the robot can have at each of those time points.  
%Furthermore, these beliefs can be computed sequentially 
%For $0\leq i\leq |\traj|$ and $0\leq j\leq i$, we use $\B_{t_i}^{t_j}$ to denote the state belief after applying control law $\tilde{\vu}_i$, at time $t_i$, assuming that 
%the last time when the localization has been active was after applying control law $\tilde{\vu}_j$, at time point $t_{j}$. That means, 
%the last time that the robot was perfectly localized was at time $t_j$, \ie after applying control law $\tilde{\vu}_j$. 
Initially, the belief is $\B_{t_0} = \delta(\vx_{t_0} - \tilde{\vx}_0)$. For $1 \leq i \leq |\traj|$ and action $\loc_{t_{i-1}}$ given according to $\sched(\B_{t_{i-1}})$ for the time window $[t_{i-1},t_i)$, the (expected) belief at time $t_i$ is:  
\begin{equation}\label{eq:seqBelief}
	\B_{t_i} =
	\begin{cases}
		%\delta(\vx_{t_i} - \tilde{\vx}_i) & \text{if } \loc_{t^-_i}=\loc_\on, \\
		\pdist(\vx_{t_i} \mid \B_{t_{i-1}}, \, \tilde{\vu}_i, \, \loc_\on) & \text{if } \loc_{t^-_i} = \loc_\on,\\
		\pdist(\vx_{t_i} \mid \B_{t_{i-1}}, \, \tilde{\vu}_i, \, \loc_\off) & \text{if } \loc_{t^-_i} \neq \loc_\on,
	\end{cases}
\end{equation}
where $\loc_{t^-_i}$ is the final status of the localization in $[t_{i-1},t_i)$.  Therefore, a discrete \textit{belief graph}, which is a directed graph that captures all possible beliefs of the robot at the localization decision points, can be constructed as in Fig.~\ref{fig:beliefs}.  The nodes of the graph are the beliefs $\B_{t_i}$, at which the localization schedule is called, and each edge is labeled with a localization action and directs to the next belief.

It is important to note that, in addition to preserving the history of collisions, $\B_{t_i}$ is dependent on the history of the applied actions.  In other words, $\B_{t_i}$ is unique to the sequence of applied localization actions up to $t_i$.  For simplicity of presentation, we do not project this history in the notation of beliefs, but it is reflected in the graph in Fig.~\ref{fig:beliefs} by only one incoming edge for each belief node.  
% Therefore, the history dependence of the beliefs causes a state explosion in the belief graph.  
This results in a state explosion in the belief graph. 
That is, the total number of nodes is exponential in the length of $\traj$, \ie $\bigO(|A|^{|\traj|})$.  We drastically reduce this size (to quadratic in $|\traj|$) by distinguishing between the collision-free and in-collision parts of the belief nodes as explained below.

% The belief graph also captures the probabilities of collision with an obstacle and ending in target, though, implicitly.  They are embedded in each of the nodes, \ie $\B_{t_i}$ is the distribution of $\vx_{t_i}$ over state space $X$, which includes obstacles $X_O$. In particular, note that in~\eqref{eq:seqBelief}, if $\loc_{t_i^-}=\loc_\on$, the control law $\tilde{\vu}_i$ is triggered only when the waypoint $\tilde{\vx}_i$ is reached, unless an obstacle is hit. Hence, in this case, the support of the belief $\B_{t_i}$ can only contain states in $X_O$ and the waypoint $\tilde{\vx}_i$. To represent the probabilities explicitly, we refine the graph by distinguishing the collision-free and in-collision part of the belief node as explained below.

\subsubsection{MDP Construction} \label{sec:mdpconst}

Recall that the robot converges to $\tilde{\B}_i$ when $a_{t_{i-1}^-} = a_\on$.  This is trivially conditioned on the fact that the robot's trajectory is collision-free.  Let $\Bof_{t_i}$ denote the collision-free part of $\B_{t_i}$, \ie the truncated probability distribution of $\vx_{t_i}$ over $X - X_O$: 
\begin{equation}
	\label{eq:beliefCF}
	\Bof_t = \pdist(\vx_t \mid \vx_{t_0}, \, \vu_{t_0:t}, \, \loc_{t_0:t}, \, \vx_t \nin X_O).
\end{equation}
% Assuming that the robot terminates after collision, the evolution of $\Bof_{t_i}$ is given in the form of \eqref{eq:seqBelief} with the conditional probability density function $\pdist$ defined only over $\vx_{t_i}\nin X_O$. 
%\begin{equation}\label{eq:seqBeliefCF}
%	\Bof_{t_i} =
%	\begin{cases}
%		%\delta(\vx_{t_i} - \tilde{\vx}_i) & \text{if } \loc_{t^-_i}=\loc_\on, \\
%		\pdist(\vx_{t_i}\nin X_O \mid \Bof_{t_{i-1}}, \, \tilde{\vu}_i, \, \loc_\on) & \text{if } \loc_{t^-_i} = \loc_\on,\\
%		\pdist(\vx_{t_i} \nin X_O \mid \Bof_{t_{i-1}}, \, \tilde{\vu}_i, \, \loc_\off) & \text{if } \loc_{t^-_i} \neq \loc_\on,
%	\end{cases}
%\end{equation}
%$\Bof_{t_i} = \pdist(\vx_{t_i} \nin X_O \mid \Bof_{t_{i-1}}, \, \tilde{\vu}_i, \, \loc_\off)$ if $\loc_{t^-_i} \neq \loc_\on$. 
% In particular, if $\loc_{t^-_i} = \loc_\on$, the collision-free part of the belief is $\Bof_{t_i} = \delta(\vx_{t_i}-\tilde{\vx}_{t_i})$. 
Then, $\Bof_{t_i} = \tilde{\B}_i$ (up to precision $\veps$) when $\loc_{t^-_i} = \loc_\on$. 
This means that, every time the localization is turned on, the truncated collision-free belief $\Bof_{t_i}$ of the robot becomes a pre-computed distribution, resulting in an independence from the history of the applied localization actions unlike $\B_{t_i}$.  
% Unlike $\B_{t_i}$, this holds independently from the sequence of localization actions applied in the past.  In other words, 
This leads to a lower number of unique $\Bof_{t_i}$ beliefs (\ie lower number of unique nodes in the graph).  
Let $\Bof^j_i$, $0\leq j\leq i$, denote the collision-free belief at time $t_i$ with the most recent localization action $a_\on$ at time $t_j$, \ie $a_{t_j^-}=a_\on$ and $a_t\neq a_\on$ for all $t_j< t< t_i$.  The sequential evolution of the collision-free beliefs becomes:
\begin{equation}\label{eq:seqBeliefCF}
\begin{cases}
\begin{aligned}
	\Bof_i^i & = \tilde{\B}_{i} & \text{if } \loc_{t^-_i} = \loc_\on,\\	
	\Bof_{i}^{j} & = \pdist(\vx_{t_i} \mid \Bof_{i-1}^{j}, \, \tilde{\vu}_i, \, \loc_\off, \, \vx_{t_i} \nin X_O) & \text{if } \loc_{t^-_i} \neq \loc_\on.
\end{aligned}
\end{cases}
\end{equation}
% \begin{equation}\label{eq:seqBeliefCF}
% 	\Bof_{t_i}^{t_i} =
% 	\begin{cases}
% 		\delta(\vx_{t_i}-\tilde{\vx}_{t_i}) & \text{if } \loc_{t^-_i} = \loc_\on,\\	
% 		\pdist(\vx_{t_i} \nin X_O \mid \Bof_{t_{i-1}}^{t_j}, \, \tilde{\vu}_i, \, \loc_\off) & \text{if } \loc_{t^-_i} \neq \loc_\on.
% 	\end{cases}
% \end{equation}
 %In particular, $\Bof_{t_i}^{t_i} = \delta(\vx_{t_i}-\tilde{\vx}_{t_i})$. 

\noindent Unlike \eqref{eq:seqBelief}, the above evolution is not deterministic but is probabilistic. That is, under each localization action, there is a probability associated with the transition from one collision-free belief to the next one, and the remaining probability mass is assigned to the collision with an obstacle. Therefore, by reasoning over $\Bof_{i}^{j}$ instead of $\B_{t_i}$, the belief graph can be greatly reduced in size at the cost of introducing probabilistic transitions.  This probabilistic model is, in fact, an MDP defined as follows. %known as a Markov decision process (MDP). 

An MDP is a tuple $\mdp = (\mdps,s_\init,\mdpa,\mdpp)$, where $\mdps$ is a non-empty finite set of states, $s_\init\in \mdps$ is the initial state, $\mdpa$ is a non-empty finite set of actions, and $\mdpp: \mdps\times \mdpa\to \distr(\mdps)$ is a (partial) probabilistic transition function. A \emph{cost function} for MDP $\mdp$ is a (partial) function $\mdpc: \mdps\times \mdpa\to \real_{\geq 0}$ such that $\mdpc(s,a)$ is defined iff 
%if and only if 
$\mdpp(s,a)$ is defined.%, for all $s\in \mdps,a\in \mdpa$.
%\end{definition}

The construction of the MDP for the evolution of the robot 
%follows the belief evolution and 
% is summarized in Algorithm~\ref{alg:mdp}. 
%Intuitively, we proceed as follows. 
is as follows. 
% The state space $\mdps$ has a specific (triangle-like) structure as shown in Fig.~\ref{fig:mdpstates}.
The state space $\mdps$ consists of 
states of the form $s_i^j$, for $0\leq i\leq |\traj|$, $0\leq j\leq i$, that correspond to beliefs $\Bof_{i}^{j}$, where $s_i^i$ indicates that the robot's belief is $\tilde{\B}_i$, and $s_0^0$ is the initial state.  In addition, $\mdps$ includes
states $s_\coll, s_\targ, s_\free$, which correspond to the robot being in collision, target, and free space, respectively. The set of MDP actions is $\mdpa=\{\loc_\off,\loc_\on,\loc_\sbo\}$, where $\loc_\sbo$ represents the sequence of actions needed to fully activate localization, which begins with $\loc_\start$, continues with $\loc_\boot$, and ends with $\loc_\on$. 

The transition probabilities %and resource consumption costs 
for states $s_i^j$ are computed as follows. 
%generated independently for each $0\leq j<|\traj|$, \ie in the diagonal levels of the triangle in Fig.~\ref{fig:mdpstates}. For a fixed $j$, transitions of states $s_i^j$ are computed sequentially, for an increasing $j\leq i<|\traj|$. 
For actions $\loc_\on$ and $\loc_\off$, the values can be computed by evolving $\Bof_i^j$ according to~\eqref{eq:seqBeliefCF}.  
% In Algorithm~\ref{alg:mdp}, subroutine $\evolve$ performs this computation and outputs the next collision-free belief along with its transition probability and total energy needed for the collision-free evolution.
In practice, techniques such as Kalman Filter or Particle Filter can be employed to compute these evolutions as well as their corresponding transition probabilities.  
% MDP cost $C$ at state $s_i^j$ under $\loc_\on$ or $\loc_\off$ is the expected value of resource usage cost $c$ from $\Bof_i^j$ to the next one under the corresponding localization action.
% With slight abuse of notation, we extend the notation in~\eqref{eq:seqBeliefCF} to $[\cdot,p,e] = \pdist(\cdot,\cdot,\cdot)$ to refer to the procedure that computes the new belief together with the probability $p$ with which the robot does not collide during the evolution and the expected total energy $e$ needed for the collision-free evolution. 
% The design of this computation can be system-dependent. For example, if the system's motion under the control law is \maja{[?]}, the values can be computed exactly, otherwise they can be approximated using techniques such as Particle Filter~\cite{particle_filter}. 
%are defined directly by the procedure $\evolve$, see Algorithm~\ref{alg:mdp}. Note that application of action $a_\off$ in a state $s_{i,j}$ can result, with non-zero probability, either in reaching state $s_{i,j+1}$ or state $s_\coll$, while $a_\on$ can only be applied in states of the form $s_{i,i}$ and results in reaching state $s_{i+1,i+1}$ with probability 1. 
% For action $\loc_\sbo$ in a state $s_i^j$, we first find the index $m$ such that $\tilde{\vx}_{m}$ would be the first waypoint reached by the system with localization system on if the booting is initialized in $\Bof_{i}^{j}$. According to~\eqref{eq:termrule}, control laws are time-triggered while localization is not yet active. Therefore, $m$ is the first index for which $\sum_{k=j+1}^{m}\Delta t_{k}\geq T_\boot$. 
For action $\loc_\sbo$ in state $s_i^j$, \ie $\Bof_{i}^{j}$, we first find the smallest index $m$, $i<m\leq |\traj|$, such that $\sum_{k=i+1}^{m}\Delta t_{k} > T_\boot$, which indicates that booting process becomes complete at some point between $t_{m-1}$ and $t_m$ if initialized at $t_i$.  Therefore, we compute the transition probabilities of $\loc_\sbo$ by combining the previously computed probabilities of action $\loc_\off$ in states $s_{i}^{j}, s_{i+1}^{j}, \ldots ,s_{m-2}^{j}$ followed by the evolution of $\Bof_{m-1}^{j}$, initialized with the localization off. After the remainder of the booting time, the localization comes on and $\Bof_m^m$ is reached.  
% The cost of $\loc_\sbo$ then becomes a direct derivation from these beliefs, resource cost $c$, and transition probabilities.
Once the computations for all $s_i^j$ states are complete, we check the probability of being in the target region or in the free space for states with $i = |\traj|$. 

The MDP cost $C$ at state $s_i^j$ under $\loc \in \mdpa$ is the expected resource usage by the robot starting from belief $\Bof_i^j$ at time point $t_i$ to the next time point $t_{i+1}$ under the corresponding localization action.  Formally,
\begin{equation} \label{eq:mdpCost}
	C(s_i^j,\loc) = \int_{t_i}^{t_{i+1}} \expv_X(c(\vx_t,\vu_t,\loc_t)) \, dt,
\end{equation}
where $\expv_X(c(\vx_t,\vu_t,\loc_t))$ is given by \eqref{eq:expcostatt}, $\vx_{t_i} \sim \Bof_i^j$, and $\loc_t \in A$ corresponds to the MDP action $a \in \mdpa$.

% by evaluating the integral of the final belief state $\B_{i,|\traj|}$ over target region $W_G$. 
% We label this transition with action $\loc_\off$ and assign energy cost of zero. %For visualization of the state space of the MDP and its transitions, see Fig.~\ref{fig:mdpstates}.

% In Algorithm~\ref{alg:mdp}, subroutine $\evolve$ is called at most twice for each state $s_i^j$, and there are $(|\traj|+2)(|\traj|+1)/2$ of these states. 
The size of the state space of this MDP is quadratic in the length of $\traj$, \ie $|\mdps| = (|\traj|+2)(|\traj|+1)/2 + 3$, and there are at most two actions per state.  The implementation of the algorithm can be made parallel by constructing the diagonal levels of the triangle-shaped state space in Fig. \ref{fig:mdpstates} independently, 
%for $0\leq j\leq |\traj|-1$, 
further reducing the complexity from quadratic to linear in the length of $\traj$.  

%%%%%%%%%%%%%%%%%%%%%%%%%%%%%%%%%%%%%%%%%%%%%%%%%%%%%%%%%%%%%%%
%%%%%%%%%%%%%%%%%%%%%%%%%%%%%%%%%%%%%%%%%%%%%%%%%%%%%%%%%%%%%%%

\subsection{Problem Reduction}\label{subsec:problemreduction}

A \emph{policy} for MDP $\mdp$ is a function $\pi:\mdps \to \distr(\mdpa)$ that associates every state with a distribution according to which the next action is chosen.  
%We denote the set of all policies in $\mdp$ by $\Pi$.
%Note that thanks to the specific form of the state space of the MDP in Algorithm~\ref{alg:mdp}, where every state is visited at most once, all policies for the MDP can be defined as $\pi:S\to \distr(A)$. 
From the above construction of the MDP $\mdp$ and definition of its action $\loc_\sbo$, it follows that every feasible localization schedule for the robot corresponds to a policy for $\mdp$, and vice versa,  every policy $\pi$ of $\mdp$ implies a feasible localization schedule $\sched_\pi$ defined as $\sched_\pi(\B_{t_i})=\pi(s_i^j)$.   Thus, $E_\cost$, $P_\coll$, and $P_\targ$ in \eqref{eq:expcosttotal}, \eqref{eq:obsprob}, and \eqref{eq:tarprob}, respectively, for the robot become the following over $\mdp$:
% The objectives for the robotics system in Problem~\ref{probform} become the following three objectives on $\mdp$:
\begin{eqnarray}
	E_\cost(\sched_\pi) &=& \expv_\pi \big( \sum_{k=0}^{|\traj|-1} \mdpc(s_k,\pi(s_k)) \big), \label{eq:mdpE} \\
	P_\coll(\sched_\pi) &=& \reach_\pi(s_\coll), \label{eq:mdpC}\\
	P_\targ(\sched_\pi) &=& \reach_\pi(s_\targ), \label{eq:mdpT}
\end{eqnarray}
where $\expv_\pi$ and $\reach_\pi(s)$ denote the expectation over the paths of $\mdp$ and the probability of reaching state $s$ under policy $\pi$, respectively.  Therefore, Problem~\ref{probform} reduces to a multi-objective optimization problem over $\mdp$, where the goal is to construct a policy that minimizes \eqref{eq:mdpE} and \eqref{eq:mdpC} and maximizes \eqref{eq:mdpT}.  

There exist algorithms and off-the-shelf tools that solve the above multi-objective problem for MDPs, \eg \cite{prism}.  These algorithms construct the Pareto front, \ie set of all optimal trade-offs between the objectives through a value iteration procedure \cite{multiomdps_atva12}.    Given a Pareto point in this set, the algorithms generate the corresponding policy using linear programming \cite{multiomdps_tacas07,FKN+11}.  The complexity of this algorithm is polynomial in the size of the MDP.  We first construct the Pareto front using these algorithms.  Then, by the choice of the designer, we generate the desired policy, \ie localization schedule.

We note that the obtained localization schedule for the continuous system with trajectory $\traj$ is optimal with respect to the user-provided waypoints, a discretization of $\traj$.  As this discretization is refined (increasing the number of waypoints), the obtained optimality approaches the true one in the continuous domain.

\section{Experimental Results}\label{sec:cs}
\newcommand{\op}{\mathsf{op}}
\newcommand{\na}{\mathsf{na}}
\newcommand{\wi}{\mathsf{wi}}
\newcommand{\en}{\mathsf{en}}
\newcommand{\eloc}{\mathsf{enl}}
\newcommand{\emot}{\mathsf{enm}}
\newcommand{\etime}{\mathsf{dur}}

In this section, we demonstrate the efficacy of the method in three case studies.  First, we illustrate the proposed framework on a scenario involving a second-order unicycle robot with energy as a resource cost and analyze the optimal performance-energy trade-offs for three different reference trajectories in simulations. 
Second, we perform the same analysis on a commercially available planetary rover and deploy the rover with the obtained schedules in an experimental setup.
Third, we demonstrate the potential of the framework in aiding the designer to choose the hardware components, specifically a computer, for a robotic system.

%%%%%%%%%%%%%%%%%%%%%%%%%%%%%%%%%%%%%%%%%%%%%%%%%%%%%%%%%%%%%
%%%%%%%%%%%%%%%%%%%%%%%%%%%%%%%%%%%%%%%%%%%%%%%%%%%%%%%%%%%%%

\subsection{Second-Order Unicycle}\label{subsec:unicycle}

\subsubsection*{Setup}
We considered a mobile robot with (second-order) unicycle dynamics given by
\begin{align}\label{eq:unicycle}
	&\dot{x}_1 = v \cos\theta + w_1, &\quad &\dot{x}_2 = v \sin\theta + w_2, \\
	&\dot{v} = u_1 + w_3, &\quad &\dot{\theta} = u_2 + w_4,
\end{align}
% where $x,y \in [0,10]$ indicate the position, and $v_x,v_y$ is the speed of the robot along the two axes. The control inputs $u_1$ and $u_2$ are the acceleration along $x$ and $y$, respectively, and the plant noise $\vw$ is sampled from a Gaussian distribution with zero mean and a diagonal covariance matrix with diagonal $(0.0001,0.0001,0.0001,0.0001)$. 
where $x_1,x_2 \in [0,10]$ indicate the position, $v$ is the speed, and $\theta$ is the heading angle of the robot. The control inputs $u_1$ and $u_2$ are the acceleration and angular velocity, respectively,
and the motion noise $\vw$ is sampled from $\mathcal{N}(\mathbf{0},\sigma^2_w \mathbf{I})$ with $\sigma_w = 0.01$, and $\mathbf{I}$ is the identity matrix. 
The robot measurements were modeled as $\vz = \vx + \vv^\star$, where sensor noise $\vv^\star \sim \N(\mathbf{0},\sigma^2_\star \mathbf{I})$ for $\star \in \{ \od, \lo\}$.  For odometry, $\sigma_\od = 0.2$, and for localization $\sigma_\lo = 0.03$.
% the noise $\vv=\vv^\od$ had a covariance matrix with diagonal $(0.04,0.04,0.04,0.04)$ and for localization sensors, the noise $\vv=\vv^\lo$ had a covariance matrix with diagonal $(0.001,0.001,0.001,0.001)$. 

We focused on trade-off analysis of the performance objectives, \ie collision avoidance and target reaching, and energy consumption as the resource objective. The energy consumption model was taken from~\cite{ondruska_icra15}. Intuitively, the energy consumed by the robot is the sum of the energy consumed by the localization module and the rest of the system. Without loss of generality, we attribute the latter mainly to the motion, \ie motors, and the CPU. Formally, the energy cost function $c_\en:X\times U\times A\to \mathbb{R}_{\geq 0}$ is defined as $c_\en=c_\eloc+c_\emot$. The localization energy cost (per unit time) $c_\eloc:X\times U\times A\to \mathbb{R}_{\geq 0}$ is:
\begin{equation} \label{eq:eloc}
	c_\eloc(\vx,\vu,a) = \begin{cases}
		\frac{\mathsf{E}_\boot}{T_\boot} & \text{if }a \in \{a_\start,a_\boot\},\\
		\mathsf{P}_\on & \text{if }a=a_\on,\\
		0 & \text{if }a=a_\off,
	\end{cases}
\end{equation}
for all $\vx\in X, \vu\in U$, where $\mathsf{E}_\boot$ and $T_\boot$ are the energy and time required to turn on the localization, respectively.  $\mathsf{P}_\on$ is its power demand when active that can be estimated as the sum of the power demand of the sensors of the localization module, taken directly from their specification, and the power consumed by the CPU for the localization algorithm that processes sensory inputs. %The latter can be obtained by first analyzing the algorithm for the number of computation instructions. Then, by using standard power-efficiency metrics, such as energy-per-instruction (EPI) or MIPS-per-watt, the number of instructions can be translated into energy consumed by the processor~\cite{EPIIntel06}. In turn, $E_L$ captures power as well as computational demand of the localization module.
The energy cost function for the rest of the system is defined as $c_\emot(\vx,\vu,a)=\mathsf{P}_{\mathsf{mot}}+\mathsf{P}_{\mathsf{CPU}}$, for all $\vx\in X, \vu\in U, a\in A$, where $\mathsf{P}_{\mathsf{mot}}$ and $\mathsf{P}_{\mathsf{CPU}}$ are power demands of the motors and the CPU (when not executing the localization algorithm). The expected energy consumption $E_\en$ is then computed using equations in Sec.~\ref{sec:energy}. In this example, we used the above model with the following parameters. 
% Intuitively, for a given state of the robot, control input and a localization action, 
The localization module used a camera and an algorithm to process the images at rate $16$\,Hz. The power demand of the module when active was $\mathsf{P}_\on=8$\,W, and the module required time $T_\boot=5$\,s and energy $\mathsf{E}_\boot=40$\,J to boot. The remaining power consumption for the robot was approximated as $\mathsf{P}_{\mathsf{mot}}+\mathsf{P}_{\mathsf{CPU}}=42$\,W. 

\begin{figure}[t]
	\begin{center}
		\includegraphics[width=.70\columnwidth]{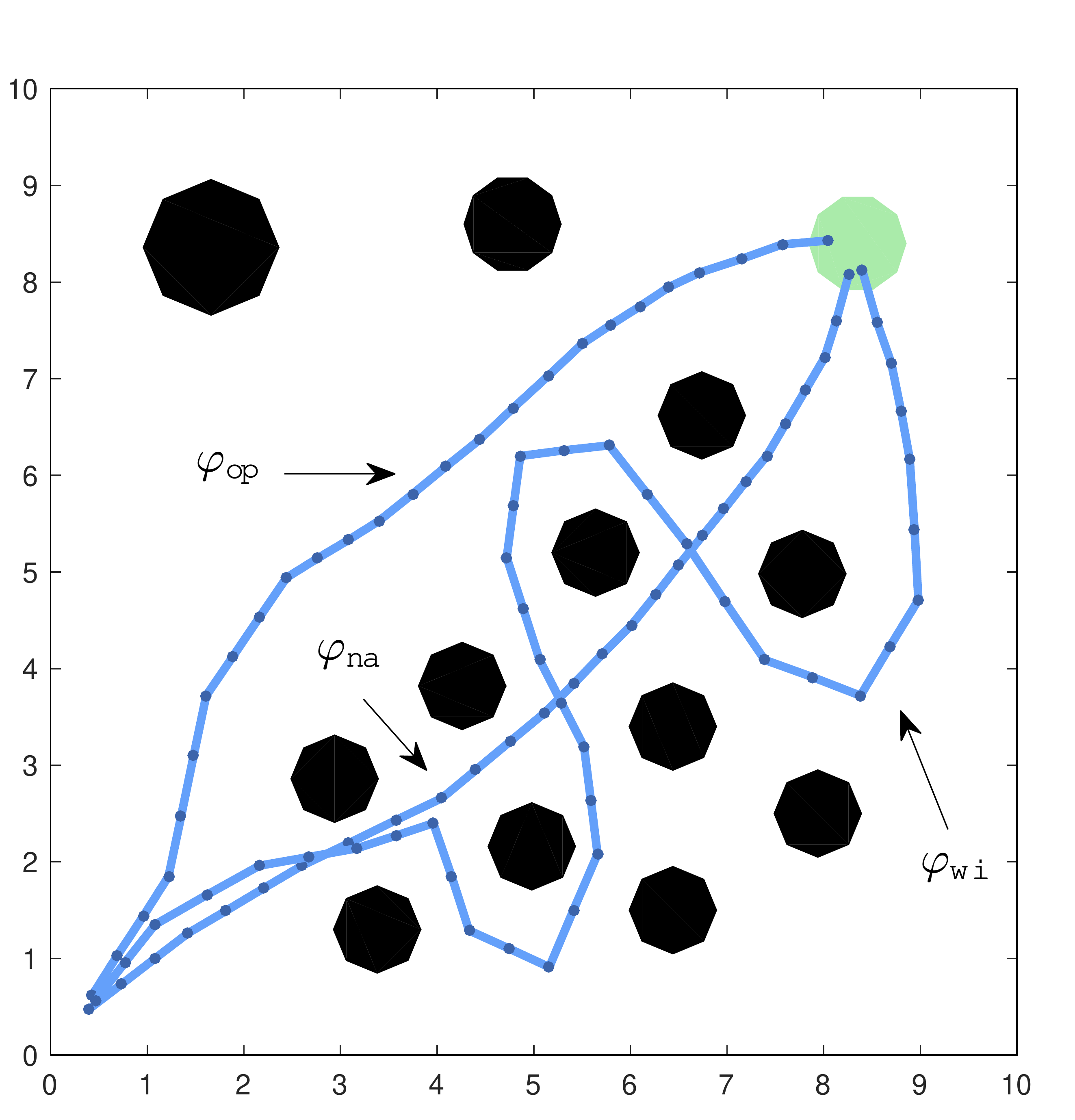}
	\end{center}
	\caption{Workspace of the unicycle robot. Obstacle and target regions are shown in black and green, respectively.  The waypoints of reference trajectories $\traj_\op$, $\traj_\na$, and $\traj_\wi$ are depicted as dark blue dots. In light blue, we show the corresponding robot trajectories using feedback controller in~\eqref{eq:controller} and localization schedule $\sched_\on$.}
\label{fig:unicycle}
\end{figure}

We considered a workspace with several obstacles and a target region, and three reference trajectories as depicted in Fig.~\ref{fig:unicycle}.  The trajectories were: 
$\traj_\op$ through open space with 26 waypoints; 
$\traj_\na$ between narrow obstacles with 27 waypoints; 
$\traj_\wi$ winding around obstacles with 39 waypoints. 
The waypoints are shown as dark blue dots in Fig.~\ref{fig:unicycle}. 
For reachability of the waypoint $\tilde{\vx} = (\tilde{x}_1,\tilde{x}_2)$ and stabilization of the belief for localization on, we used dynamic feedback linearization (DFL) controller to linearize the unicycle dynamics as in \cite{oriolo2002wmr} and an LQG controller on the linearized dynamics given by: 
\begin{eqnarray} \label{eq:controller}
	% \begin{aligned}
    &u_1 = \bar{u}_1 \cos \theta + \bar{u}_2 \sin \theta, \quad
	u_2 = \frac{\bar{u}_2}{v} \cos \theta -\frac{\bar{u}_1}{v} \sin \theta, &\nonumber \\ 
% 	\begin{aligned}
% 		u_1 &= \bar{u}_1 \cos \theta + \bar{u}_2 \sin \theta, \\
% 		u_2 &= \frac{\bar{u}_2}{v} \cos \theta -\frac{\bar{u}_1}{v} \sin \theta,
% 	\end{aligned}
		&\bar{u}_1 = k_1 (\tilde{x}_1 - \hat{x}_1) - k_2 \, \hat{x}_3 \cos \hat{x}_4,&\\ 
		&\bar{u}_2 = k_3 (\tilde{x}_2 - \hat{x}_2) - k_4 \, \hat{x}_3 \sin \hat{x}_4,&\nonumber
	% \end{aligned}
\end{eqnarray}
where $\hat{x}_i$ is the $i$th component of the mean of the state estimate, the output of Kalman filter, and $k_1 = k_3 = 1$ and $k_2 = k_4 = 2.236$ are feedback gains. 
Fig.~\ref{fig:unicycle} depicts the robot trajectories under no noise.
% that follow the reference trajectories using the controller \eqref{eq:controller} and localization schedule $\sched_\on$, which keeps the localization module on at all times.

\begin{table*}[ht]
\caption{Pareto fronts computed for the second-order unicycle. For each of the three reference trajectories, we list the corner points of the Pareto front, namely their probability of reaching the target $P_\targ$, the probability of collision $P_\coll$ and the expected total energy $E_\en$ with the fraction of it consumed by the localization system $E_\eloc$. For comparison, we also list the performance guarantees of the schedule $\sched_\on$ and the percentage of the total and localization energy saved by Pareto-optimal schedules compared to $\sched_\on$. On the right, Pareto fronts are visualized in three-dimensional space. For better readability of the images, we plot the projection of the polytopes onto two planes.
}\label{tab:unicycle}
\begin{center}
%\vspace*{0.5cm}
{\small (a) Open trajectory $\traj_\op$.}\\
%\vspace*{0.3cm}
\begin{minipage}{.99\textwidth}
\centering
{\scriptsize
\renewcommand{\arraystretch}{1.5}
\begin{tabular}{ | l | S[table-format=1.4] S[table-format=1.4] S[table-format=5.2] S[table-format=5.2] | S[table-format=5.2] S[table-format=5.2] |}
\cline{2-5}
\multicolumn{1}{l |}{} & \multicolumn{1}{c}{$P_\targ$} & \multicolumn{1}{c}{$P_\coll$} & \multicolumn{1}{c}{$E_\en$} & \multicolumn{1}{c |}{fraction $E_\eloc$} & & \multicolumn{1}{c}{} \\
\cline{1-5}
$\sched_\on$ & 1 & 0 & 21790.20 & 3486.44 & & \multicolumn{1}{c}{} \\
\hline
Pareto points: & & & & & \multicolumn{2}{|c|}{$E_\en$ and $E_\eloc$ saved over $\sched_\on$} \\
\hline
1 ($\sched_\off$) & 0.8640 & 0 & 4502.19 & 0 & 79.34\% & 100.00\% \\
2 & 1 & 0 & 5243.21 & 176.91 & 75.94\% & 94.93\% \\
\hline
\end{tabular}
}
\hspace*{0.3cm}
\raisebox{-0.5\height}{\includegraphics[width=.32\linewidth]{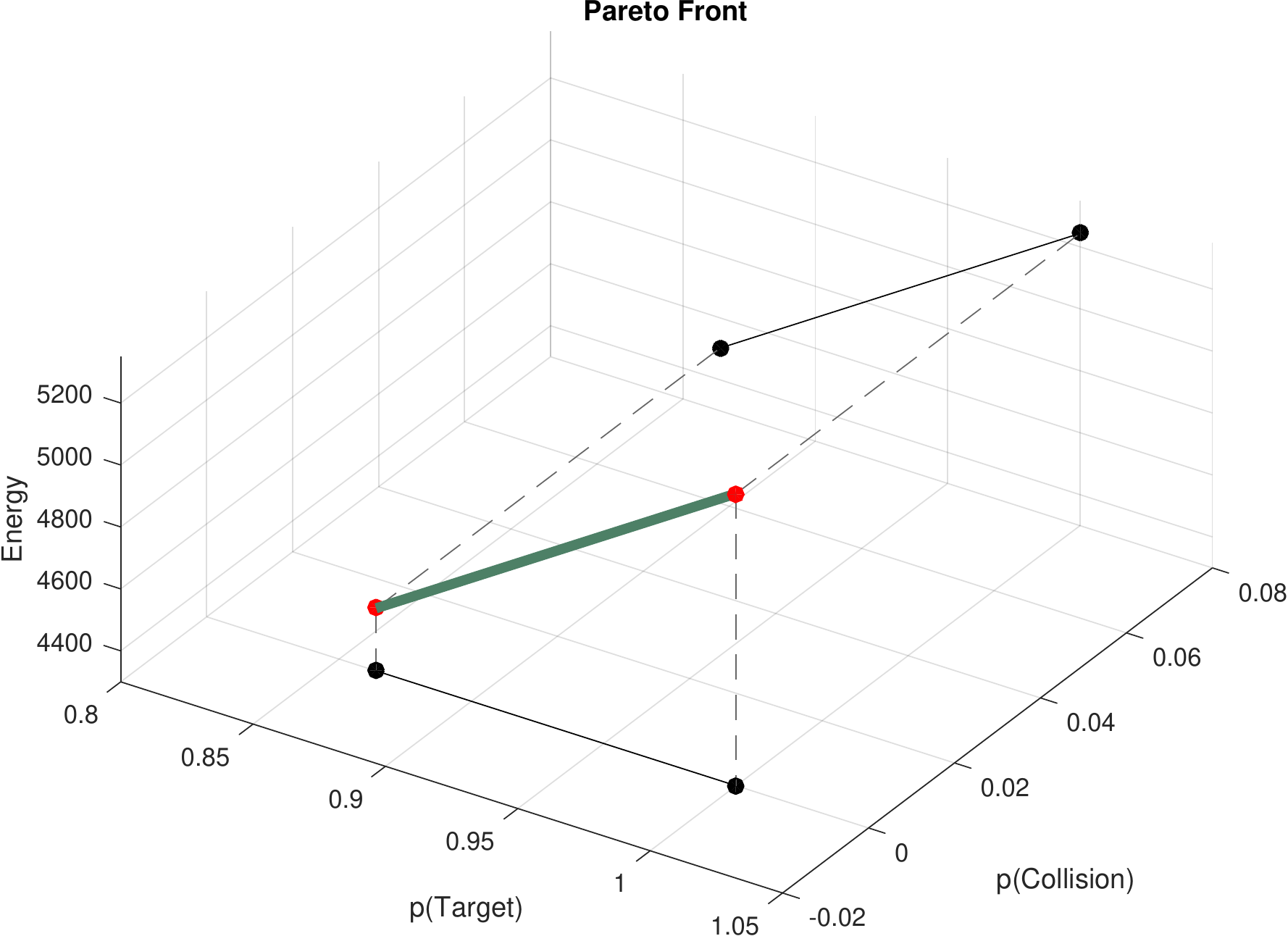}}
\end{minipage}
\vspace*{0.3cm}\\
{\small (b) Narrow trajectory $\traj_\na$.}\\
\vspace*{0.2cm}
\begin{minipage}{.99\textwidth}
\centering
{\scriptsize
\renewcommand{\arraystretch}{1.5}
\begin{tabular}{ | l | S[table-format=1.4] S[table-format=1.4] S[table-format=5.2] S[table-format=5.2] | S[table-format=5.2] S[table-format=5.2] |}
\cline{2-5}
\multicolumn{1}{l |}{} & \multicolumn{1}{c}{$P_\targ$} & \multicolumn{1}{c}{$P_\coll$} & \multicolumn{1}{c}{$E_\en$} & \multicolumn{1}{c |}{fraction $E_\eloc$} & & \multicolumn{1}{c}{} \\
\cline{1-5}
$\sched_\on$ & 1 & 0 & 22476.90 & 3596.30 & & \multicolumn{1}{c}{} \\
\hline
Pareto points: & & & & & \multicolumn{2}{|c|}{$E_\en$ and $E_\eloc$ saved over $\sched_\on$} \\
\hline
1 & 0.8540 & 0 & 8213.72 & 840.07 & 63.46\% & 76.64\% \\
2 & 1 & 0 & 8935.53 & 1013.56 & 60.25\% & 71.82\% \\
3 & 0.9980 & 0.0020 & 8211.67 & 869.09 & 63.47\% & 75.83\% \\
4 & 0.8523 & 0.0020 & 7491.30 & 695.95 & 66.67\% & 80.65\% \\
5 & 0.9940 & 0.0060 & 7459.44 & 690.57 & 66.81\% & 80.80\% \\
6 & 0.8250 & 0.0060 & 6734.18 & 516.92 & 70.04\% & 85.63\% \\
7 & 0.9841 & 0.0159 & 6748.25 & 519.60 & 69.98\% & 85.55\% \\
8 & 0.9703 & 0.0297 & 5959.24 & 337.93 & 73.49\% & 90.60\% \\
9 & 0.8054 & 0.0297 & 5290.72 & 174.74 & 76.46\% & 95.14\% \\
10 & 0.8306 & 0.0159 & 6018.15 & 346.52 & 73.23\% & 90.36\% \\
11 & 0.9131 & 0.0869 & 5060.09 & 159.01 & 77.49\% & 95.58\% \\
12 ($\sched_\off$) & 0.7689 & 0.0869 & 4396.48 & 0.00 & 80.44\% & 100.00\% \\
\hline
\end{tabular}
}
\hspace*{0.3cm}
\raisebox{-0.5\height}{\includegraphics[width=.32\linewidth]{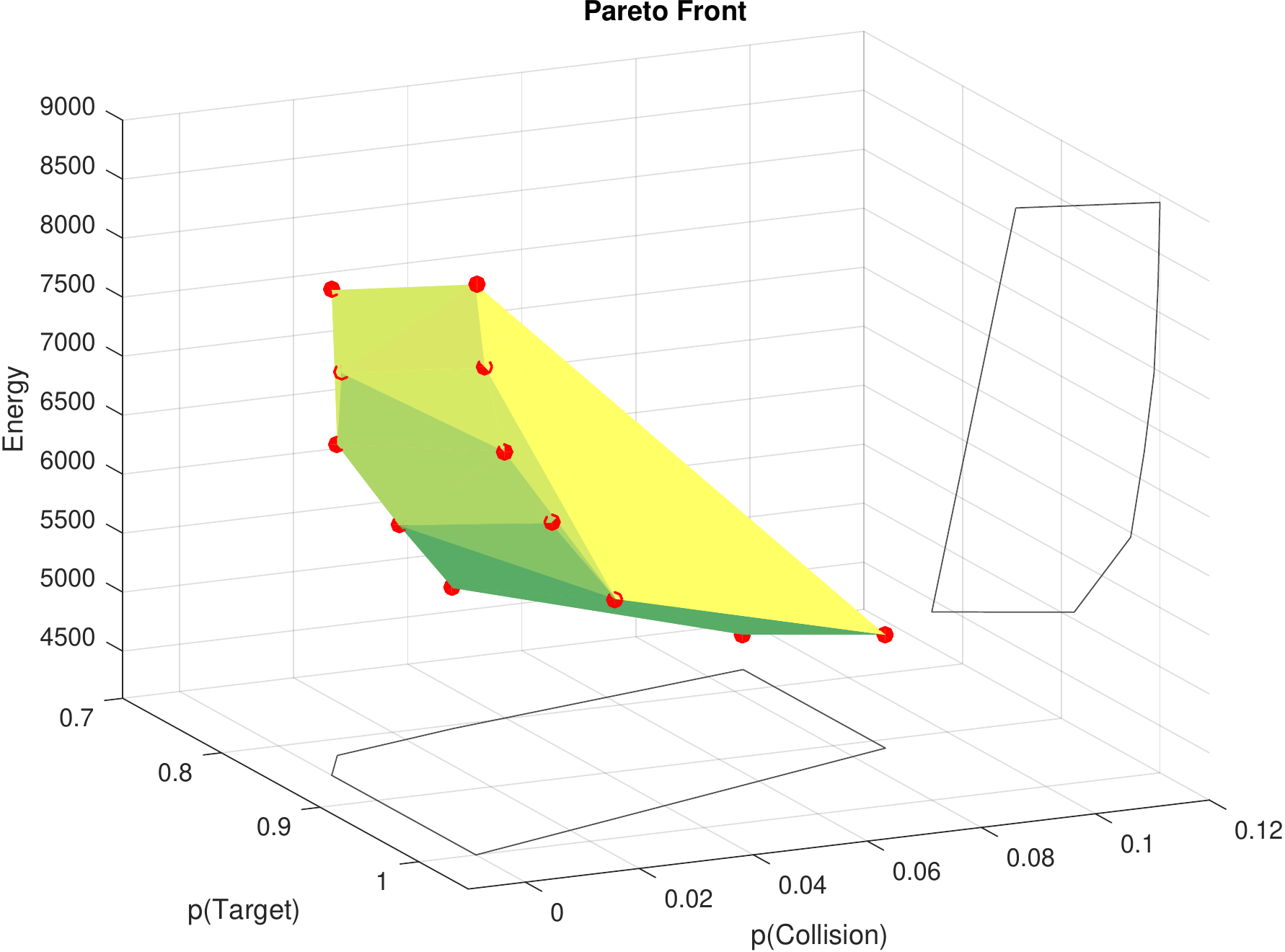}}
\end{minipage}
\vspace*{0.4cm}\\
{\small (c) Winding trajectory $\traj_\wi$.}\\
\vspace*{0.2cm}
\begin{minipage}{.99\textwidth}
\centering
{\scriptsize
\renewcommand{\arraystretch}{1.5}
\begin{tabular}{ | l | S[table-format=1.4] S[table-format=1.4] S[table-format=5.2] S[table-format=5.2] | S[table-format=5.2] S[table-format=5.2] |}
\cline{2-5}
\multicolumn{1}{l |}{} & \multicolumn{1}{c}{$P_\targ$} & \multicolumn{1}{c}{$P_\coll$} & \multicolumn{1}{c}{$E_\en$} & \multicolumn{1}{c |}{fraction $E_\eloc$} & & \multicolumn{1}{c}{} \\
\cline{1-5}
$\sched_\on$ & 1 & 0 & 33354.40 & 5336.70 & & \multicolumn{1}{c}{} \\
\hline
Pareto points: & & & & & \multicolumn{2}{|c|}{$E_\en$ and $E_\eloc$ saved over $\sched_\on$} \\
\hline
1 & 0.9120 & 0 & 10596.90 & 837.37 & 68.23\% & 84.31\% \\
2 & 0.9200 & 0 & 10634.30 & 811.57 & 68.12\% & 84.79\% \\
3 & 1 & 0 & 11316.77 & 1011.48 & 66.07\% & 81.05\% \\
4 & 0.9980 & 0.0020 & 10633.27 & 876.41 & 68.12\% & 83.58\% \\
5 & 0.9102 & 0.0020 & 9914.84 & 702.64 & 70.27\% & 86.83\% \\
6 & 0.9261 & 0.0020 & 9977.65 & 674.78 & 70.09\% & 87.36\% \\
7 & 0.9243 & 0.0040 & 9296.83 & 540.38 & 72.13\% & 89.87\% \\
8 & 0.9940 & 0.0060 & 9297.63 & 539.62 & 72.12\% & 89.89\% \\
9 & 0.9006 & 0.0060 & 8560.97 & 363.32 & 74.33\% & 93.19\% \\
10 & 0.9683 & 0.0317 & 8325.83 & 340.42 & 75.04\% & 93.62\% \\
11 & 0.8773 & 0.0317 & 7608.22 & 168.68 & 77.19\% & 96.84\% \\
12 & 0.9012 & 0.0988 & 7294.80 & 157.48 & 78.13\% & 97.05\% \\
13 ($\sched_\off$) & 0.8345 & 0.0988 & 6642.58 & 0 & 80.08\% & 100.00\% \\
14 & 0.9012 & 0.0433 & 7640.69 & 179.72 & 77.09\% & 96.63\% \\
\hline
\end{tabular}
}
\hspace*{0.3cm}
\raisebox{-0.5\height}{\includegraphics[width=.32\linewidth]{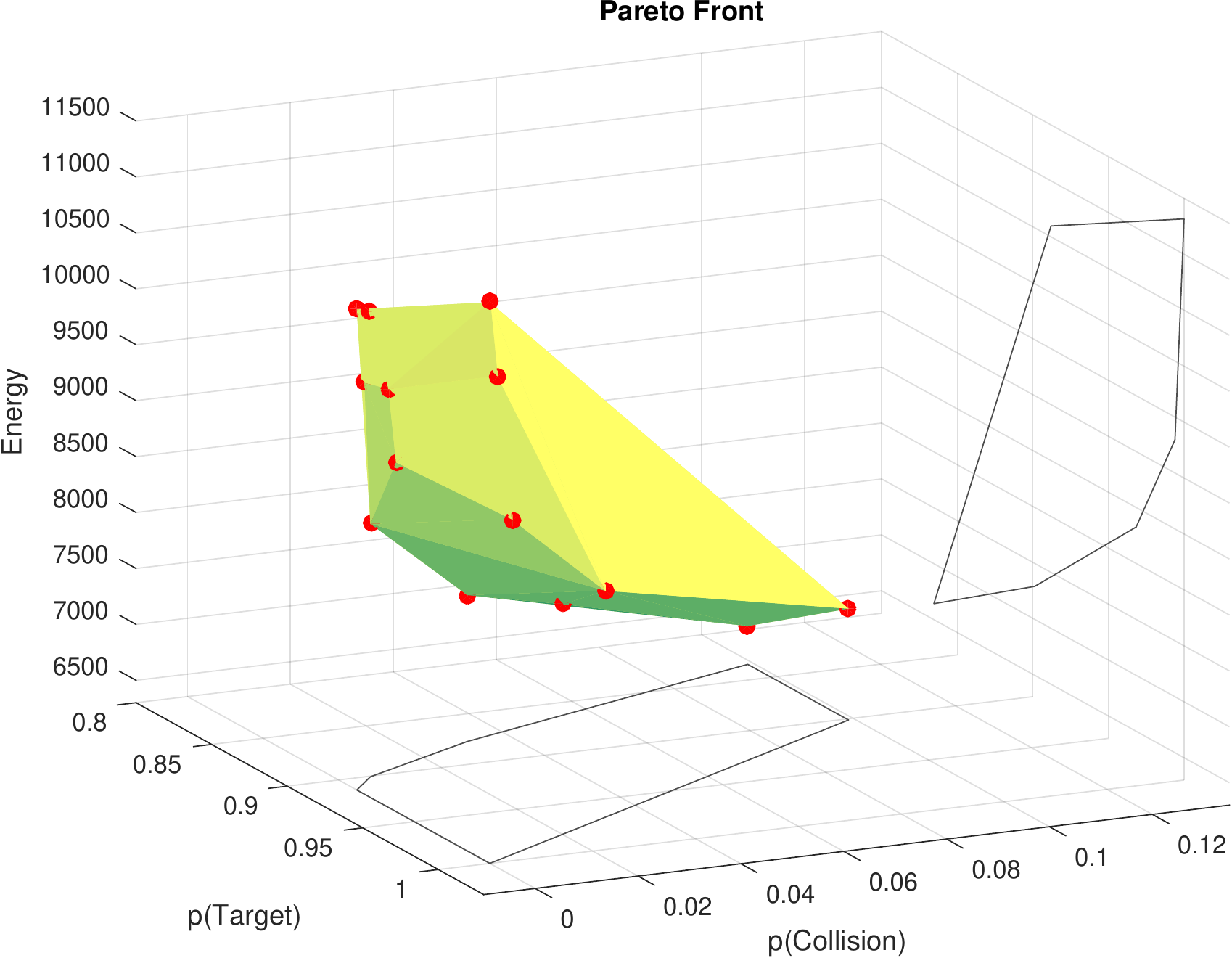}}
\end{minipage}
\end{center}
\end{table*}

\begin{figure*}
	\begin{center}
		\small
		\begin{tabular}{c c c}
			\includegraphics[height=.2\linewidth]{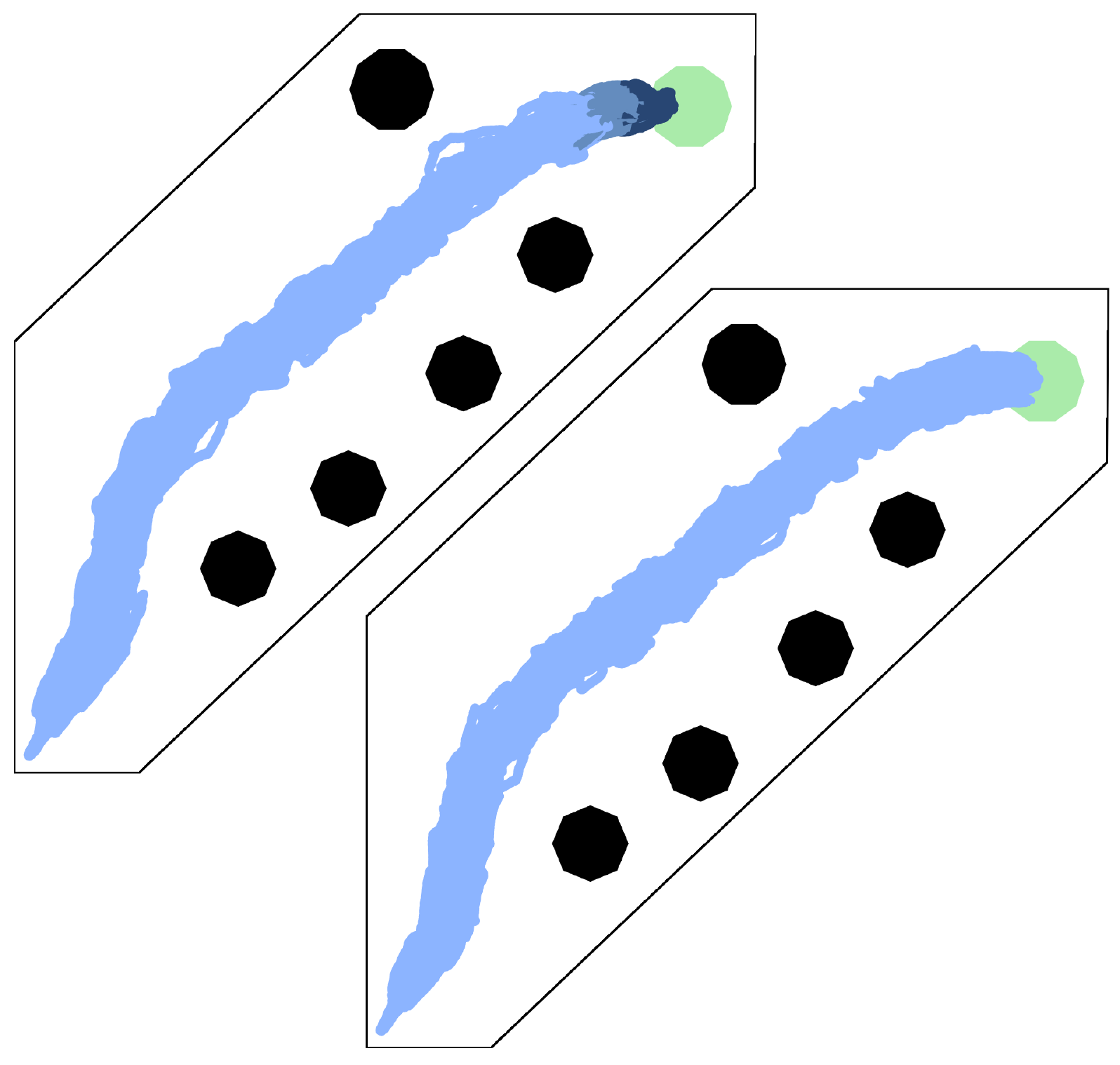} \quad \quad &
			\includegraphics[height=.2\linewidth]{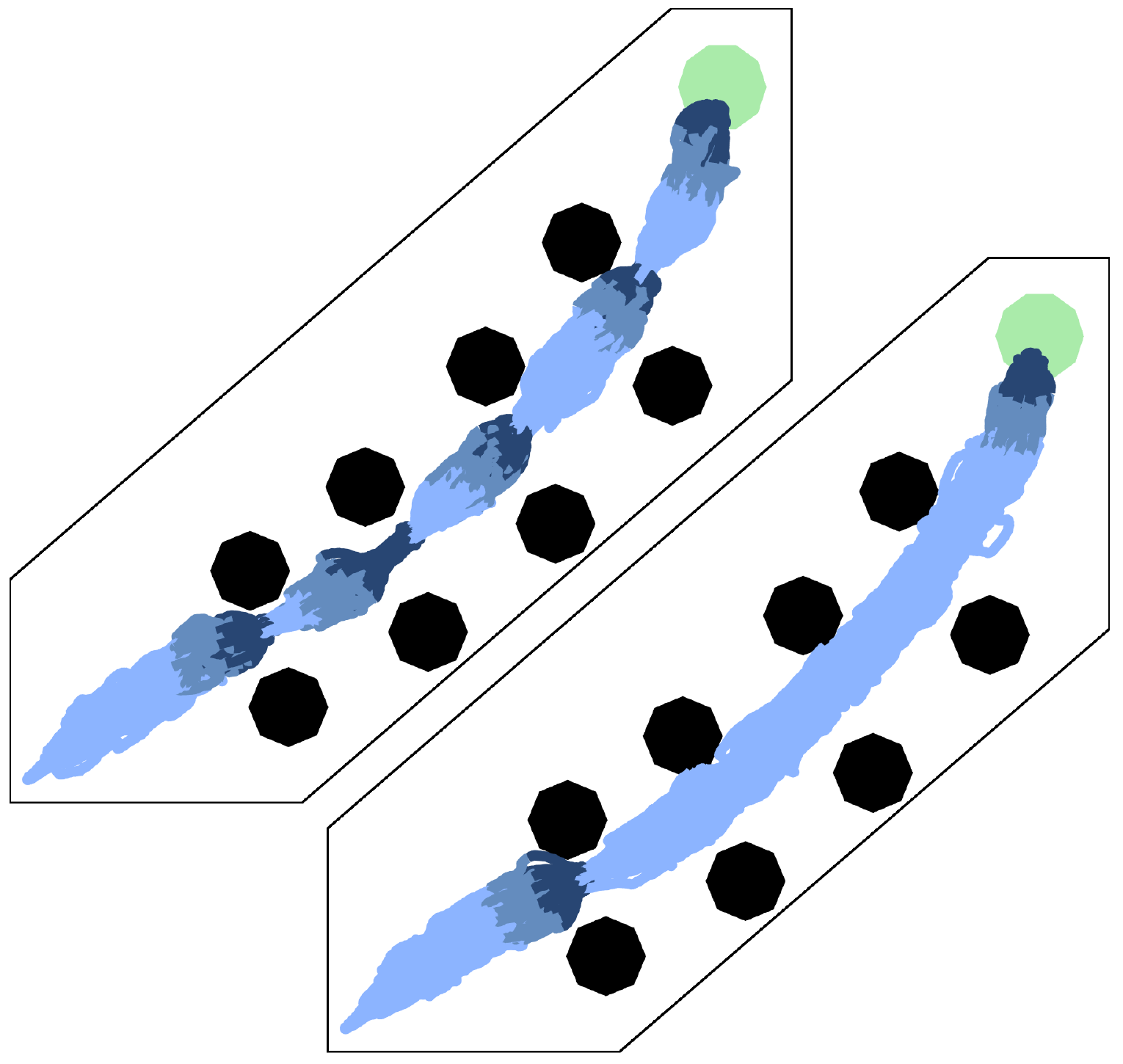} & \quad \quad
			\includegraphics[height=.2\linewidth]{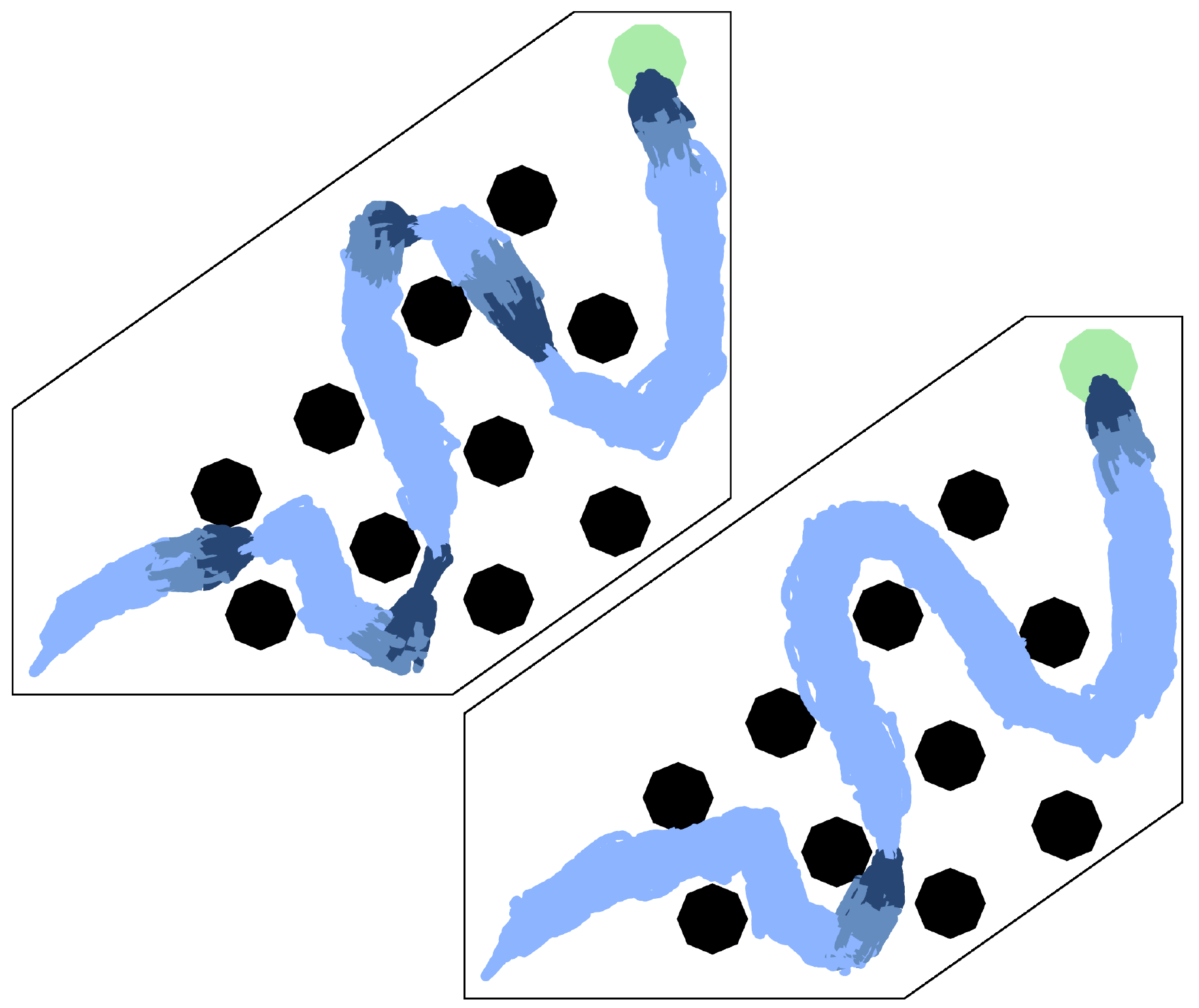}\\
			& & \\
			(a) Open trajectory $\traj_\op$. \quad \quad &
			(b) Narrow trajectory $\traj_\na$. & \quad \quad
			(c) Winding trajectory $\traj_\wi$.
		\end{tabular}
	\end{center}
\caption{Sample trajectories under two localization schedules for each trajectory $\traj_\op$, $\traj_\na$, and $\traj_\na$ for the unicycle robot.  The light, medium, and dark blue trajectory segments indicate localization status $a_\off$, $a_\start$ and $a_\boot$, and $a_\on$, respectively. In the top-left images in (a)-(c), the performance guarantees are $P_\targ = 1$ and $P_\coll = 0$, the same as $\sched_\on$, while saving 60\% to 75\% energy over $\sched_\on$.  In the bottom-right images, the performance guarantees are $P_\targ = 0.86$ and $P_\coll = 0$ for $\traj_\op$ and $P_\targ = 0.97$ and $P_\coll = 0.03$ for $\traj_\na$ and $\traj_\wi$ in return for additional energy saving of 73\% to 75\%.}
\label{fig:unicyclesim}
\end{figure*}

\subsubsection*{Pareto Fronts}
We constructed an MDP for each reference trajectory according to the abstraction algorithm in Sec.~\ref{subsec:abstraction} by using a Particle Filter.  They consisted of 656, 708, and 1\,488 state-action pairs for $\traj_\op$, $\traj_\na$, and $\traj_\wi$ respectively. We then computed the Pareto front for the three objectives of collision avoidance, target reaching and energy consumption for each reference trajectory using PRISM-games~\cite{prism-games}. In Table \ref{tab:unicycle}, we present the plots and list the vertices of the convex Pareto fronts and compare their values against the ones of the schedule $\sched_\on$, which keeps the localization module on at all times.  

Every point on the surface of the Pareto front corresponds to a particular optimal trade-off between the three objectives and there exists a localization schedule that achieves it. A bound on a value of an objective, \eg the probability of collision should be at most $0.01$, imposes a slice through the surface that divides the Pareto front into trade-offs which are achievable with the desired bound and those which are not. A projection of the Pareto front to a lower dimension is the Pareto front for the subset of objectives. For example, in plots in Tables \ref{tab:unicycle}a-c, the projection onto the bottom plane represents the optimal trade-offs between the target-reaching and collision probabilities, regardless of the expected energy consumption.

In all three cases, the localization schedule $\sched_\on$ is not Pareto optimal. That means it is not efficient to keep localization on at all times because there exist localization schedules that have the same probabilistic guarantees as $\sched_\on$, \ie $P_\targ = 1$ and $P_\coll = 0$, while saving energy by turning localization off. On the other hand, the localization schedule $\sched_\off$, which keeps the localization off at all times, is Pareto optimal in all three cases. While $\sched_\off$ tends to decrease the energy consumption, this may not be a desirable schedule due to low performance guarantees. Generally, Pareto-optimal schedules save 60-80\% of total energy and 72-100\% of localization energy compared to $\sched_\on$.

\subsubsection*{Localization Schedules}

For each reference trajectory, we generated schedules for two Pareto points using PRISM~\cite{prism}.  We simulated 500 sample robot trajectories under each localization schedule and Fig.~\ref{fig:unicyclesim} depicts 100 of them.  The first set of schedules correspond to the Pareto points with $P_\targ = 1$, $P_\coll = 0$ (top-left images in Fig.~\ref{fig:unicyclesim}).  As shown in these figures, to ensure that the target region is reached with probability 1, these schedule turn on localization only at the critical waypoints: near obstacles to ensure safety and right before the target to ensure ending in target.  The second set of schedules correspond to $\sched_\off$ for $\traj_\op$ and Pareto points with $P_\targ = 0.97$ and $P_\coll = 0.03$ for $\traj_\na$ and $\traj_\wi$ (bottom-right images in Fig.~\ref{fig:unicyclesim}).  These schedules trade off the performance by a small percentage to save energy.  As shown in bottom-right image in Fig. \ref{fig:unicyclesim}a, this schedule keeps localization off for the entire $\traj_\op$, resulting in missing the target 14\% of times (loss in performance) in trade-off for 79\% gain in energy. For $\traj_\na$ and $\traj_\wi$, the schedules turn on the localization only at two extremely critical points; one very close to an obstacle and one before the target.  These schedules trade off 3\% loss in performance to gain 73\% to 75\% in energy.  
To summarize, the simulations suggest that the need for the localization to be active grows with the level of noise and proximity of obstacles.

\subsubsection*{Validation}
In the simulations above, the average performance of the robot with respect to all three objectives was within 3\% of the theoretical values (performance guarantees of the Pareto points).  In addition, we randomly selected 10 more Pareto points and performed similar computations.  All the simulation results were within 4\% of the theoretical values.  We note that these error values are expected to decrease as the number of simulations and the number of particles in the particle filter increase.

%%%%%%%%%%%%%%%%%%%%%%%%%%%%%%%%%%%%%%%%%%%%%%%%%%%%%%%%%%%%%
%%%%%%%%%%%%%%%%%%%%%%%%%%%%%%%%%%%%%%%%%%%%%%%%%%%%%%%%%%%%%

\subsection{Rover Experiments}\label{subsec:rover}

\subsubsection*{Setup}
The robotic platform used in this experimental case study is ARC Q14 planetary rover shown in Fig.~\ref{fig:Resources}.  It is designed to mimic the configuration and specification found on rovers deployed for planetary exploration.  The rover's base is rectangular (0.8 $m$ by 0.9 $m$) and has 4 wheels and 8 motors.  It can operate in two kinematic modes: Ackermann steering and differential drive with maximum speed of 0.5 $m/s$.  The robot is equipped with a Point Grey Bumblebee XB3 camera. 
We use Dub4 \cite{Dub4} as the high accuracy localisation module, while low accuracy measurements are obtained using Visual Odometry \cite{VO}.
% The localization module (high accuracy measurements) is based on visual teach and repeat \cite{FurgaleJFR10}, and low accuracy measurements are based on visual odometry.  
The on-board computations are carried out on MicroSVR computer.  The energy consumption model of the robot is the same as the one in Sec.~\ref{subsec:unicycle} taken from \cite{oriolo2002wmr} that previously studied this platform.  

% \subsubsection*{Setup}
We modeled the motion of the rover as the unicycle in Sec.~\ref{subsec:unicycle} with constrained velocity, turn angle, and acceleration.  We used the same DFL as above to linearize the dynamics and employed receding horizon controller for reachability.  Kalman filter was utilized for state estimation.  The online control computations were performed in MATLAB on a MacBook Pro with 2.7 GHz Intel Core i5 and 8 GB of memory, which communicated to the robot via Wi-Fi.  We estimated  motion and measurement noise as $\N(\mathbf{0},\sigma^2 \mathbf{I})$, where $\sigma_w = 0.1$, $\sigma_\od = 0.1$, and $\sigma_\lo = 0.01$, and the frequency of sensor measurements was 4 Hz. The robot's task was to navigate from an entrance to exit door of a 10$m$-by-6$m$ meeting room cluttered with various furniture pieces.  The robot was first driven by a human to learn the reference trajectory $\traj$ (teach phase), during which the localization module automatically extracts waypoints of $\traj$.  The environment and these waypoints are shown in Fig.~\ref{fig:roverExp}a.

\subsubsection*{Pareto Front}
We computed the Pareto front for this scenario by first generating the abstraction MDP and then our multi-objective algorithm.  We considered the same objectives as in Sec.\ref{subsec:unicycle}; the vertices of the Pareto front are shown in Table~\ref{tab:rover}.  In this case study, both $\sched_\on$ and $\sched_\off$ are Pareto optimal; one gives rise to the highest $P_\targ$ and the other results in the smallest $E_\en$.  Note that it is possible to save 18\%, 24\%, and 32\% in $E_\en$ by sacrificing small percentage (0.5\%, 1\%, and 5\%, respectively) in $P_\targ$.  

\subsubsection*{Robot Deployment}
\begin{figure*}
	\begin{center}
		\small
		\begin{tabular}{c c c}
			\includegraphics[width=55mm]{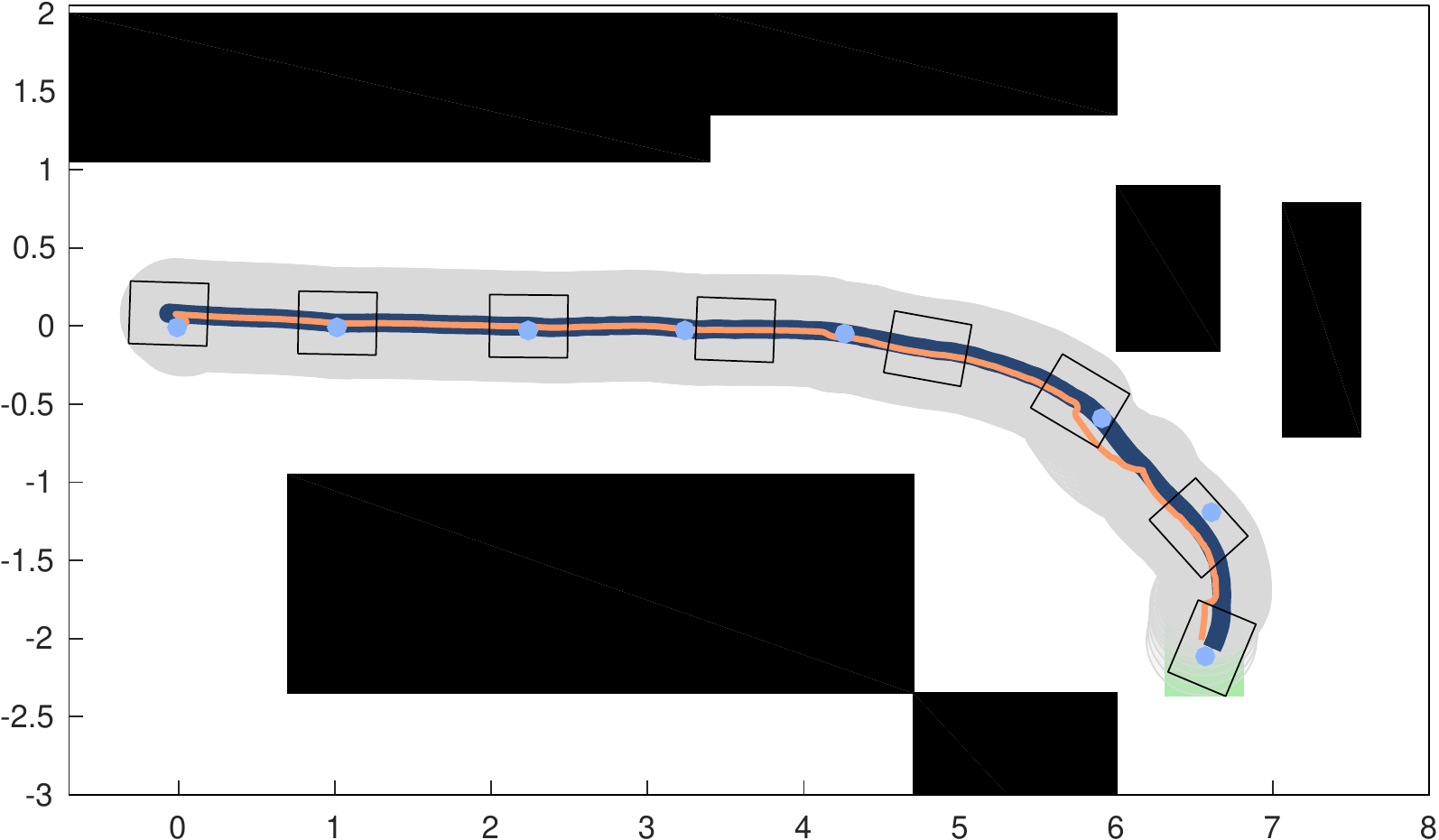} \quad &
			\includegraphics[width=55mm]{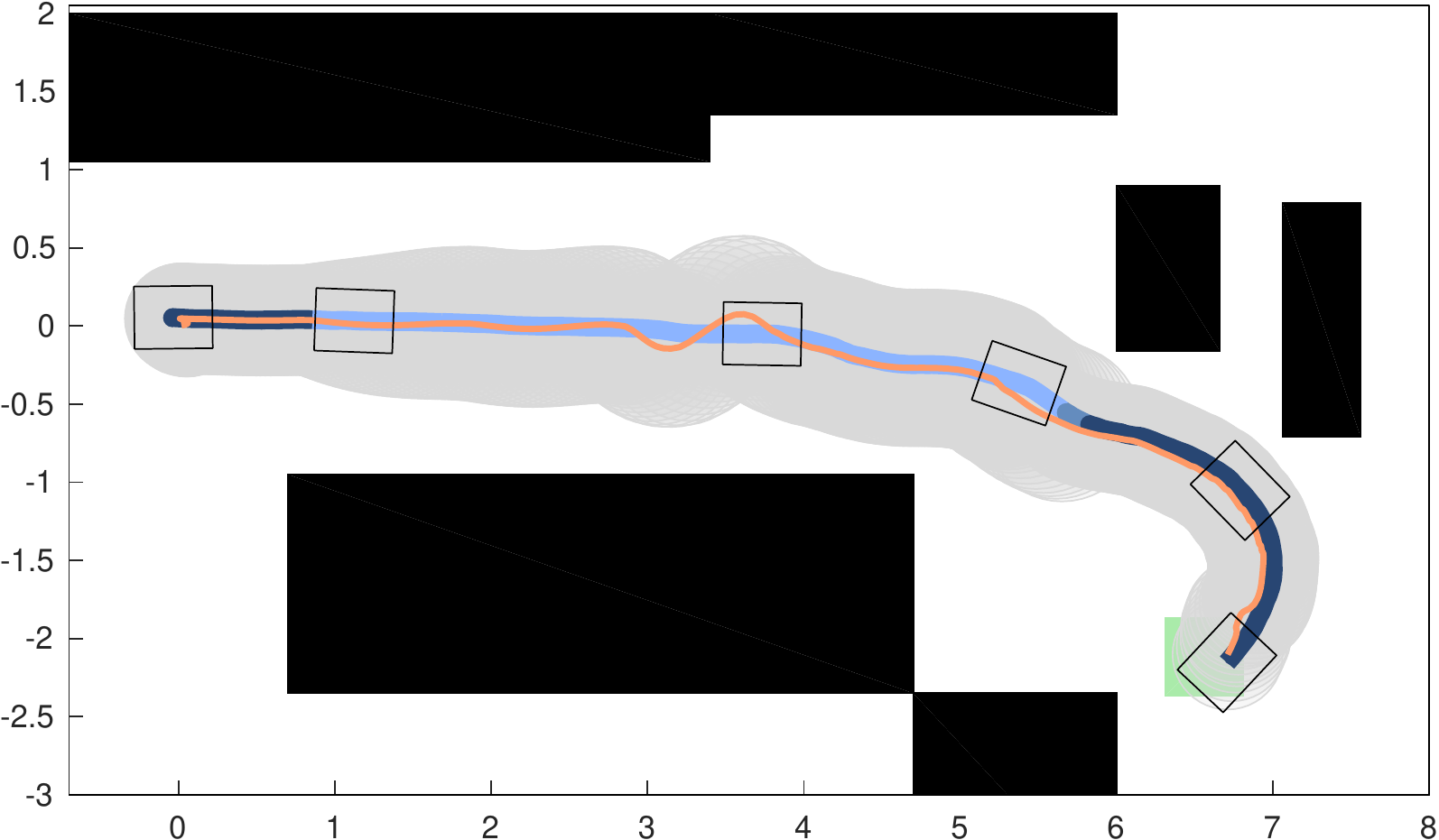} & \quad
			\includegraphics[width=55mm]{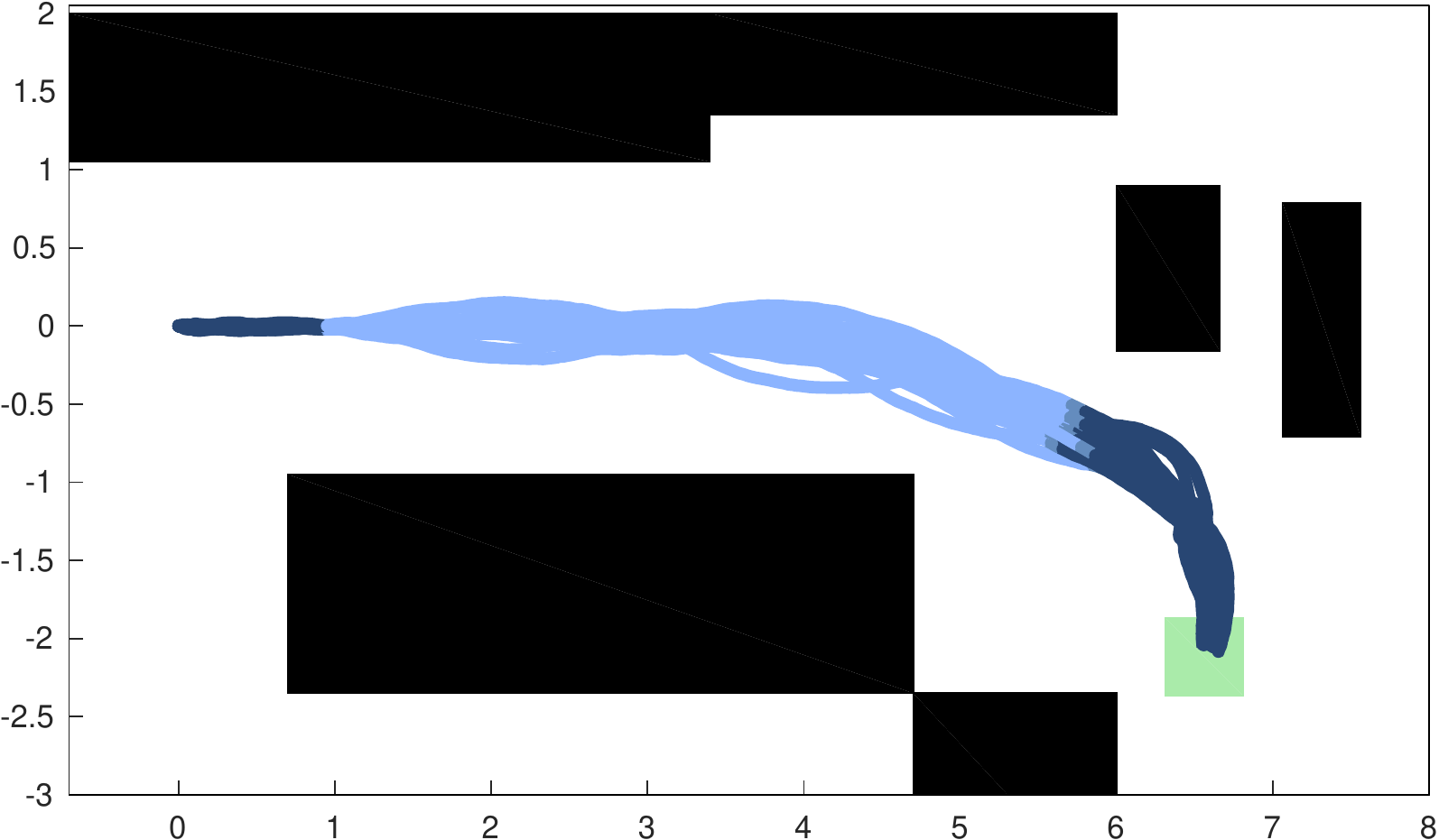}\\
			& & \\
			(a) Robot trajectory under $\sched_\on$. \quad &
			(b) Robot trajectory under $\sched_3$. & \quad
			(c) Simulation trajectories under $\sched_3$.
		\end{tabular}
	\end{center}
\caption{Robot trajectories in experiments and simulations.  The waypoints are shown as light blue dots in (a).  The light, medium, and dark blue trajectory segments indicate localization status $a_\off$, $a_\start$ and $a_\boot$, and $a_\on$, respectively. In (a) and (b), robot's belief is shown in orange (state estimate) and gray (projection of variance), and robot's orientation by black-edged boxes. Under $\sched_\on$, the performance guarantees are $P_\targ = 1$ and $P_\coll = 0$ while they are $P_\targ = 0.99$ and $P_\coll = 0.01$ for $\sched_3$ in trade-off for  24\% reduction in $E_\en$.}
\label{fig:roverExp}
\end{figure*}

We deployed the robot under $\sched_\on$ and $\sched_3$.  Fig.~\ref{fig:roverExp}b-c show the robot's trajectories, localization status in different shade of blue, state estimate in orange, and belief's variance's projection onto 2-D in gray.  The robot itself is shown as black-edged rectangles along the trajectory.  As evident in these figures, under $\sched_\on$, the robot is always safe because it is able to stay within a very close proximity of $\traj$ at all times.  Under $\sched_3$, the robot uses its localization only at the very beginning and for the last two waypoints.  The use of localization at the beginning sets the robot's trajectory and belief on the right path.  Once localization is turned off, the uncertainty in the robot's belief grows, but the robot is still able to continue with the path without deviating too far from the $\traj$ thanks to its initial localization.  Once the robot is near a point that is dangerously close to an obstacle, and $\traj$ requires sharp maneuvers, the robot turns on its localization to reduce its uncertainty and enable itself to perform the maneuvers.  Note that, under $\sched_3$, once the localization is turned back on, on account of the increased uncertainty, the robot is required to make a sharper turn than under $\sched_\on$  to be able to reach the target.
% it realizes that its orientation is a bit off due to uncertainty it accumulated during $a_\off$, so it needs to turn more sharply than it does under $\sched_\on$ in order to be able to reach the target.  
The framework is aware of such uncertainties; therefore, under $\sched_3$, the performance guarantee is reduced by $1\%$ to save 24\% in energy in comparison to $\sched_\on$, resulting in an elongation of the battery life.  Fig.~\ref{fig:roverExp}c illustrates 50 trajectories that was obtained in simulation prior to deployment of the robot.  Note that this figure shows only the trajectory of the center of the robot; the robot's volume needs to be added to every point along the trajectory.

\begin{table}
\caption{Pareto fronts for the planetary rover. $E_\en$ and $E_\eloc$ denote and the total and localization energy, respectively. 
% We only list points with $P_\targ\geq 0.95$. 
For comparison, we list the energy savings of Pareto-optimal schedules compared to $\sched_\on$.
}
\label{tab:rover}
	\begin{center}
		\begin{minipage}{.49\textwidth}
		    \centering
		    % \vspace*{0.3cm}
				{\scriptsize
				\renewcommand{\arraystretch}{1.5}
				\begin{tabular}{ | l | S[table-format=1.4] S[table-format=1.4] S[table-format=5.2] S[table-format=5.2] | S[table-format=2.2] S[table-format=2.2] |}
					% \cline{2-5}
					% \multicolumn{1}{l |}{} & \multicolumn{1}{c}{$P_\targ$} & \multicolumn{1}{c}{$P_\coll$} & \multicolumn{1}{c}{$E_\en$} & \multicolumn{1}{c |}{$E_\etime$} & & \multicolumn{1}{c}{} \\
					% \cline{1-5}
					% $\sched_\on$ & 1 & 0 & 9468.67 & 182.09 & & \multicolumn{1}{c}{} \\
					\hline
					P. Pt & \multicolumn{1}{c}{$P_\targ$} & \multicolumn{1}{c}{$P_\coll$} & \multicolumn{1}{c}{$E_\en$} & \multicolumn{1}{c |}{$E_\eloc$} & \multicolumn{2}{|c|}{$E_\en$, $E_\eloc$ saved} \\
					\hline
					1 &	0.5000 & 0.0000 & 6155.52 & 507.90 & 29.99\% & 65.34\% \\
					2 ($\sched_\on$)& 1.0000 & 0.0000 & 8791.87 & 1465.31 & 0.00\% & 0.00\% \\
					3 & 0.9900 & 0.0100 & 6713.23 & 650.88 & 23.64\% & 55.58\% \\
					4 & 0.9950 & 0.0050 & 7215.16 & 814.58 & 17.93\% & 44.41\% \\
					5 ($\sched_\off$) & 0.4428 & 0.0050 & 4805.86 & 0.00 & 45.34\% & 100.00\% \\
					6 & 0.9552 & 0.0448 & 5969.96 & 325.25 & 32.10\% & 77.80\%\\
					\hline
				\end{tabular}
				}
		\end{minipage}
	\end{center}
\end{table}

%%%%%%%%%%%%%%%%%%%%%%%%%%%%%%%%%%%%%%%%%%%%%%%%%%%%%%%%%%%%%
%%%%%%%%%%%%%%%%%%%%%%%%%%%%%%%%%%%%%%%%%%%%%%%%%%%%%%%%%%%%%

\subsection{Robot with choices of PCs}\label{sec:hardware}

Hardware choices in robot design affect the capabilities of the robot and can result in different achievable resource-performance trade-offs. In this example, we analyzed resource-performance trade-offs for a mobile robot with two different mini PCs. This type of analysis can aid the designer in choosing the best suitable hardware to achieve a desired level of performance. 

\subsubsection*{Setup}
\begin{figure}[t]
\begin{center}
\includegraphics[width=.75\columnwidth]{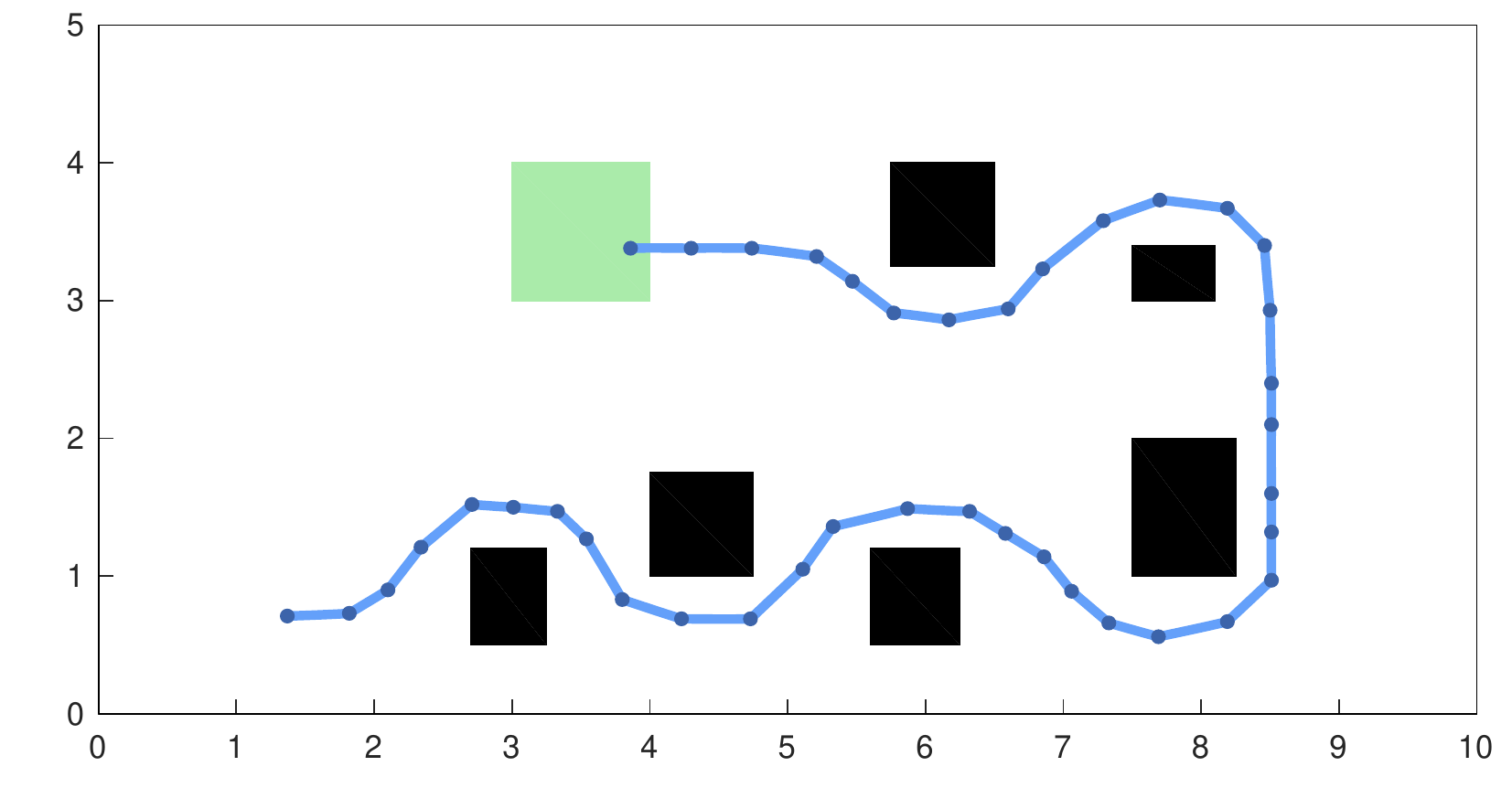}
\end{center}
\caption{Workspace and reference trajectory of the mobile robot. The waypoints of the reference trajectory are depicted as dark blue dots. In light blue, we show the corresponding robot trajectory using nominal system.}
\label{fig:hardware}
\end{figure}

We modeled the robot dynamics as
\begin{align*}
    % \dot{\vx} = A \vx + B \vu, \quad 
    %     A= \begin{pmatrix} -0.3 & 0.1\\ 0.1 & -0.3\end{pmatrix},\quad
    %     B=\begin{pmatrix} 1 & 0\\ 0 & 1\end{pmatrix},
    \dot{x}_1 &= 0.1 x_2 - 0.3 x_1 +  u_1 + w_1, \\
    \dot{x}_2 &= 0.1 x_1 - 0.3 x_2 +  u_2 + w_2
\end{align*}
where $x_1 \in [0,10]$ and $x_2 \in [0,5]$ indicate the 2-D position of the robot and the process noise distribution is $\N(\mathbf{0},0.07^2 \mathbf{I})$. The measurement models for odometry and localization module were $\vz=\vx +\vv$ with noise distribution $\N(\mathbf{0},0.2^2 \mathbf{I})$ and $\N(\mathbf{0},0.03^2 \mathbf{I})$, respectively. We considered the workspace shown in Fig.~\ref{fig:hardware} with obstacles and a target region and a trajectory with 40 waypoints. The controllers for the waypoints were designed by LQG method. When localization system was on, a controller was terminated when the state estimate reaches (a proximity of) the steady state distribution around the associated waypoint. The time triggers for localization off were time durations computed based on the nominal system, \ie without process and measurement noise.

We analyzed the robot running with IPC i3 Barebone ($\sim$\$600) % $609, i3-3227U
and IPC2 i5 Barebone ($\sim$\$700). % $706, i5-4300U
The two computers processed localization measurements at rates proportional to their computational power, namely $10$\,Hz and $16$\,Hz, respectively. With IPC2, the robot received feedback about its position at higher rate which allowed for faster convergence to waypoints. Thus using the localization module, the robot traversed the trajectory in less time with IPC2 than with IPC. Odometry measurements were processed at $20$\,Hz when the localization module was not active with both computers. We then focused on trade-off analysis for the performance objectives $P_\targ$ and $P_\coll$, and two resource objectives of the expected energy consumption $E_\en$ and expected trajectory duration $E_\etime$.

The energy consumption was given by the following parameters. 
%The localization module used a Point Grey Bumblebee XB3 camera for visual localization with power consumption $5$\,W, and power consumption of motors was estimated as $30$\,W.
The power consumption of localization module and motors were estimated as $5$\,W and $30$\,W, respectively. The IPC and IPC2 require power $17$\,W and $15$\,W, respectively. This amounts to an overall power consumption of $44$\,W and $42$\,W when localization module was active, respectively. When localization was not active, the power savings were estimated as $5$\,W for the sensors and one fifth of the computer power consumption, \ie $8$\,W for both computers. The module required time $T_\boot=5$\,s and energy $40$\,J to boot.

\subsubsection*{Pareto Fronts}
The MDP constructed for the reference trajectory consisted of 1\,529 state-action pairs for both computers. The convex Pareto fronts for the four objectives of target reaching, collision avoidance, energy consumption and trajectory duration had 11 vertices for IPC and 22 vertices for IPC2. In Table~\ref{tab:hardware}, we list only the vertices for which the probability of reaching the target $P_\targ$ is at least $0.95$. 

In neither cases, localization schedule $\sched_\on$ that keeps the localization module on at all times is Pareto optimal because there exist localization schedules that have the same probabilistic guarantees as $\sched_\on$, \ie $P_\targ = 1$ and $P_\coll = 0$, while saving energy and time by turning localization off. On the other hand, localization schedule $\sched_\off$, which keeps the localization off at all times, is Pareto optimal for both IPC and IPC2 as it minimizes the expected energy consumption and duration. This however comes at a cost of poor performance guarantees. As indicated in Table~\ref{tab:hardware}, generally, Pareto-optimal schedules save 34-60\% of energy and 31-52\% of time compared to $\sched_\on$ for IPC, and 20-50\% of energy and 17-40\% of time for IPC2. 

\subsubsection*{Hardware Comparison} 
For given performance guarantees, the robot can complete the reference trajectory faster with IPC2 than with IPC, \eg in $118.87$\,s compared to $126.42$\,s for $P_\targ=1$ and $P_\coll=0$.  More interestingly, while the IPC2 has higher power requirements with $17$\,W relative to $15$\,W for IPC, the analysis shows that the robotic system consumes less energy with IPC2 than with IPC in completing the trajectory. That means that IPC2 offers overall better resource-performance trade-offs than IPC for the given trajectory. This, however, comes at the cost of higher dollar value for IPC2. 

The designer may also wish to choose the computer based on the duration or energy rather than performance guarantees as the primary preference. For example, if it is sufficient to complete the trajectory in expected time $E_\etime\leq 130$\,s, IPC is a better choice because it can achieve the best performance at lower dollar value. On the other hand, if the constraint is $E_\etime\leq 120$\,s, only IPC2 can guarantee best performance.

We can also analyze the Pareto fronts in more detail. For example, consider time constraint $E_\etime\leq 90$\,s. By projecting the Pareto fronts onto $E_\en$ and fixing $E_\etime= 90$\,s, we obtain the convex sets of trade-offs between performance objectives $P_\targ$ and $P_\coll$ achievable by the two computers within the given time bound. For IPC, the set has vertices $P_\targ=0.938, P_\coll=0.062$ (for maximizing $P_\targ$), and $P_\targ=0.887, P_\coll=0$ (for minimizing $P_\coll$). For IPC2, the set has vertices $P_\targ=0.943, P_\coll=0.057$ (for maximizing $P_\targ$), and $P_\targ=0.897, P_\coll=0.047$ (for minimizing $P_\coll$). Thus, while finishing the trajectory within $90$\,s, there exists a localization schedule for IPC that guarantees no collision while no such schedule exists for IPC2. On the other hand, with IPC2 the achievable probability of reaching the target location is higher than for IPC. 
Using this analysis, the designer can make a well-informed choice according to given preferences.

\begin{table}
\caption{Pareto fronts for the planetary rover with two different mini PCs. In the tables, $E_\en$ and $E_\etime$ denote and the energy consumption and trajectory duration, respectively. We only list points with $P_\targ\geq 0.95$. For comparison, we list the performance guarantees of the schedule $\sched_\on$ and the energy and time savings of Pareto-optimal schedules compared to $\sched_\on$.
}
\label{tab:hardware}
\begin{center}
\begin{minipage}{.49\textwidth}
    \centering
    {\small (a) IPC i3 Barebone.}\\
    \vspace*{0.3cm}
	{\scriptsize
	\renewcommand{\arraystretch}{1.5}
	\begin{tabular}{ | l | S[table-format=1.4] S[table-format=1.4] S[table-format=5.2] S[table-format=5.2] | S[table-format=2.2] S[table-format=2.2] |}
	\cline{2-5}
	\multicolumn{1}{l |}{} & \multicolumn{1}{c}{$P_\targ$} & \multicolumn{1}{c}{$P_\coll$} & \multicolumn{1}{c}{$E_\en$} & \multicolumn{1}{c |}{$E_\etime$} & & \multicolumn{1}{c}{} \\
	\cline{1-5}
	$\sched_\on$ & 1 & 0 & 9468.67 & 182.09 & & \multicolumn{1}{c}{} \\
	\hline
	P. Pt & & & & & \multicolumn{2}{|c|}{$E_\en$, $E_\etime$ saved} \\
	\hline
	%0.9180 & 0.0000 & 6102.21 & 124.04 & 35.55\% & 31.88\%\\
	1 & 1.0000 & 0.0000 & 6282.54 & 126.42 & 33.65\% & 30.57\%\\
	2 & 0.9980 & 0.0020 & 5759.97 & 117.06 & 39.17\% & 35.71\%\\
	%0.9162 & 0.0020 & 5580.00 & 114.69 & 41.07\% & 37.02\%\\
	3 & 0.9920 & 0.0080 & 5026.35 & 104.72 & 46.92\% & 42.49\%\\
	%0.9107 & 0.0080 & 4847.46 & 102.36 & 48.81\% & 43.79\%\\
	4 & 0.9821 & 0.0179 & 4689.18 & 99.67 & 50.48\% & 45.26\%\\
	%0.9055 & 0.0179 & 4506.69 & 97.24 & 52.40\% & 46.60\%\\
	%0.9260 & 0.0354 & 4297.36 & 94.22 & 54.62\% & 48.26\%\\
	%0.9340 & 0.0660 & 3987.71 & 89.05 & 57.89\% & 51.10\%\\
	5 ($\sched_\off$) & 0.8798 & 0.0660 & 3827.27 & 86.98 & 59.58\% & 52.23\%\\
	\hline
	\end{tabular}
	}
\end{minipage}

\vspace*{0.5cm}

\begin{minipage}{.49\textwidth}
    \centering
    {\small (b) IPC2 i5 Barebone.}\\
    \vspace*{0.3cm}
	{\scriptsize
	\renewcommand{\arraystretch}{1.5}
	\begin{tabular}{ | l | S[table-format=1.4] S[table-format=1.4] S[table-format=5.2] S[table-format=5.2] | S[table-format=2.2] S[table-format=2.2] |}
	\cline{2-5}
	\multicolumn{1}{l |}{} & \multicolumn{1}{c}{$P_\targ$} & \multicolumn{1}{c}{$P_\coll$} & \multicolumn{1}{c}{$E_\en$} & \multicolumn{1}{c |}{$E_\etime$} & & \multicolumn{1}{c}{} \\
	\cline{1-5}
	$\sched_\on$ & 1 & 0 & 7185.22 & 143.71 & & \multicolumn{1}{c}{} \\
	\hline
	P. Pt & & & & & \multicolumn{2}{|c|}{$E_\en$, $E_\etime$ saved} \\
	\hline
	%0.9200 & 0.0000 & 5645.05 & 117.65 & 21.44\% & 18.13\%\\
	1 & 1.0000 & 0.0000 & 5762.75 & 118.87 & 19.80\% & 17.28\%\\
	2 & 0.9980 & 0.0020 & 5214.40 & 110.93 & 27.43\% & 22.81\%\\
	%0.9401 & 0.0020 & 5095.48 & 109.69 & 29.08\% & 23.67\%\\
	3 & 0.9960 & 0.0040 & 4985.78 & 106.74 & 30.61\% & 25.72\%\\
	%0.9382 & 0.0040 & 4867.09 & 105.50 & 32.26\% & 26.59\%\\
	4 & 0.9900 & 0.0100 & 4691.53 & 101.92 & 34.71\% & 29.08\%\\
	%0.9326 & 0.0100 & 4573.56 & 100.68 & 36.35\% & 29.94\%\\
	5 & 0.9880 & 0.0120 & 4604.97 & 100.55 & 35.91\% & 30.03\%\\
	%0.9307 & 0.0120 & 4487.23 & 99.32 & 37.55\% & 30.89\%\\
	6 & 0.9841 & 0.0159 & 4470.06 & 97.95 & 37.79\% & 31.84\%\\
	%0.9270 & 0.0159 & 4352.79 & 96.73 & 39.42\% & 32.69\%\\
	7 & 0.9802 & 0.0198 & 4371.77 & 96.33 & 39.16\% & 32.97\%\\
	%0.9233 & 0.0198 & 4254.98 & 95.10 & 40.78\% & 33.82\%\\
	8 & 0.9645 & 0.0355 & 4098.98 & 92.87 & 42.95\% & 35.37\%\\
	%0.9067 & 0.0355 & 3981.19 & 91.61 & 44.59\% & 36.25\%\\
	%0.9396 & 0.0604 & 3889.56 & 89.53 & 45.87\% & 37.70\%\\
	%0.8833 & 0.0604 & 3774.81 & 88.31 & 47.46\% & 38.55\%\\
	9 ($\sched_\off$) & 0.8570 & 0.0883 & 3601.42 & 85.75 & 49.88\% & 40.33\%\\
	%0.9117 & 0.0883 & 3719.50 & 87.07 & 48.23\% & 39.41\%\\
	%0.9473 & 0.0527 & 3949.98 & 90.87 & 45.03\% & 36.77\%\\
	%0.8924 & 0.0527 & 3815.73 & 89.26 & 46.89\% & 37.89\%\\
	\hline
	\end{tabular}
	}
\end{minipage}
\end{center}
\end{table}

\section{Final Remarks and Future Work}\label{sec:concl}
We have introduced a general framework for the exploration of performance-resource trade-offs and demonstrated its efficacy in module scheduling and robot design on case studies. The framework can be adapted to schedule other modules such as perception or different motors. 
% It can also be adapted to handle robots with noise in their motion (motors), in addition to sensors, with a simple modification to the control law rules.  That is, when the localization is on, the control law triggers as soon as the robot reaches the waypoint's neighborhood, whose radius is a function of the motion noise, instead of the exact convergence to the waypoint. 
The framework can also be extended to scenarios, in which the aim is to schedule the use of more than one localization module, providing state estimates at different levels of certainty. Intuitively, the only change to the framework is in the abstraction step, \ie the MDP construction.  The additional modules cause an increase in both action set and state space of the MDP. 

There are multiple directions for future work. First, in this work we focused on total expected resource costs, but more complex cost measures such as long-run average or ratio costs might be of interest.  
Second, the delay between making an observation and computing the corresponding state estimate introduces additional uncertainty.  An attractive future direction is to explore the Pareto front by considering these delays.
 % We intend to approach the performance-resource trade-off problem for delayed systems more rigorously. 
Finally, solving the problem for a combination of modules, \eg localization and perception, and, ultimately, localization, perception and planning, poses a great challenge, which should be explored for full trade-off analysis of the autonomy modules.

% The work has focused on total expected resource cost, 
% %which, under some simplifying assumptions (e.g. neglecting caching effects and sleep/wake cycles), allow us to obtain an approximate measure of energy per instruction that can be used to select the type of processor. 
% In future, we intend to employ more complex cost measures, e.g., long-run average or ratio costs. These have been formulated for MDPs and can be incorporated within our framework.
% %While in this work we focus on three specific objectives, \ie total energy consumption, collision avoidance and target reaching, the developed framework can be extended, in a straightforward way, to handle a wide range of objectives including temporal tasks.
% %This is due to the fact that the multi-objective optimization for MDPs has been solved and implemented for a large class of objectives and their combinations.  



% \section*{ACKNOWLEDGMENT}

%%%%%%%%%%%%%%%%%%%%%%%%%%%%%%%%%%%%%%%%%%%%%%%%%%%%%%%%%%%%%%%%%%%%%%%%%%%%%%%%

% \bibliographystyle{abbrv}
\bibliographystyle{IEEEtran}
\bibliography{references}

\end{document}